%% file: main.tex
\documentclass[10pt, a4paper, logo]{googledeepmind} %
\usepackage{times}

\usepackage{hyperref}
\usepackage[leftcaption]{sidecap} 

\input{math_commands.tex}

\usepackage{hyperref}
\usepackage{url}
\usepackage{booktabs}
\usepackage{graphicx}
\usepackage{subcaption}
\usepackage{amssymb}
\usepackage[most]{tcolorbox}
\usepackage[table]{xcolor}
\tcbuselibrary{breakable} 
\usepackage{amsmath}
\usepackage{siunitx}
\usepackage{colortbl}

\usepackage{microtype}
\usepackage{color}
\usepackage{wrapfig}
\usepackage{natbib}
\usepackage{colortbl}
\usepackage{microtype}
\usepackage{graphicx}
\usepackage{booktabs} 
\usepackage{array}
\usepackage{textcomp}
\usepackage{stfloats}
\usepackage{float}
\usepackage{verbatim}
\usepackage[ruled, linesnumbered, lined]{algorithm2e}
\usepackage{multirow}
\usepackage{enumitem}
\usepackage[table]{xcolor} 
\usepackage{tabularx}      
\newcolumntype{C}{>{\centering\arraybackslash}X}
\newcolumntype{L}{>{\raggedright\arraybackslash}X}
\newcolumntype{R}{>{\raggedleft\arraybackslash}X}
\usepackage{siunitx}       
\usepackage{siunitx}
\usepackage{booktabs}
\usepackage{colortbl}
\usepackage{xcolor}
\usepackage{siunitx}
\usepackage{pifont} 

\definecolor{PrimaryBlue}{HTML}{1E5AA8}      
\definecolor{PrimaryLight}{HTML}{E8F4FD}     
\definecolor{PrimaryDark}{HTML}{0F3A6E}      

\definecolor{SuccessGreen}{HTML}{2E8B57}     
\definecolor{SuccessLight}{HTML}{D4EDDA}     

\definecolor{WarningAmber}{HTML}{F0A020}     
\definecolor{WarningLight}{HTML}{FFF3CD}     

\definecolor{DangerCoral}{HTML}{DC5A4A}      
\definecolor{DangerLight}{HTML}{F8D7DA}      

\definecolor{NeutralGray}{HTML}{6C757D}      
\definecolor{LightGray}{HTML}{F8F9FA}        
\definecolor{BorderGray}{HTML}{DEE2E6}       

\definecolor{BestColor}{HTML}{D4EDDA}        
\definecolor{SecondBestColor}{HTML}{FFF3CD}  
\definecolor{mygray}{gray}{0.9}
\definecolor{ggg}{RGB}{46,139,87}            
\definecolor{rrr}{RGB}{220,90,74}            
\definecolor{oodc}{RGB}{30,90,168}           
\definecolor{idc}{RGB}{46,139,87}

\usepackage{tcolorbox}
\usepackage{amsmath,amsfonts}

\def\Bias#1#2{\bm{b}}

\newtcolorbox{examplebox}[2][]{ 
    breakable, 
    enhanced, 
    colback=white, 
    colframe=cyan, 
    coltitle=white, 
    fonttitle=\bfseries, 
    title=#2, 
    overlay middle={\draw[cyan, line width=1pt](frame.south west)--(frame.south east);}, 
    overlay last={\draw[cyan, line width=1pt](frame.south west)--(frame.south east);}, 
    #1 
}

\newtcolorbox{acmbluebox}[1]{%
    enhanced,
    colback=PrimaryLight,
    colframe=PrimaryBlue!70,
    boxrule=0.8pt,
    arc=3pt,
    left=8pt,right=8pt,top=6pt,bottom=6pt,
    title={#1},
    coltitle=white,
    colbacktitle=PrimaryBlue!70,
    fonttitle=\bfseries\small\sffamily,
    attach boxed title to top left={xshift=6mm,yshift=-3mm},
    boxed title style={
        sharp corners,
        boxrule=0pt,
        arc=2pt,
        left=6pt,right=6pt
    },
    shadow={1pt}{-1pt}{0pt}{black!15},
}

\usepackage[T1]{fontenc}
\usepackage{booktabs}      
\usepackage{graphicx}      
\usepackage[table]{xcolor} 
\usepackage{siunitx}       
\usepackage{etoolbox}      
\usepackage[normalem]{ulem}     

\definecolor{impcolor}{HTML}{2E8B57} 

\newcommand{\improvementstyle}[1]{$^{\textcolor{impcolor}{\tiny #1}}$}

\newcommand{\scoreimp}[2]{%
  \textbf{#1}%
  \ifstrequal{#2}{+0.0}{}{%
    \ifstrequal{#2}{0.0}{}{%
      \makebox[0pt][l]{\improvementstyle{#2}}%
    }%
  }%
}

\title{Grounding the Score: Explicit Visual Premise Verification for Reliable VLM Process Reward Models}

\author[1,2]{Junxin Wang\textsuperscript{$\spadesuit$}\textsuperscript{*}}
\author[1]{Dai Guan\textsuperscript{$\spadesuit$}}
\author[3]{Weijie Qiu}
\author[1]{Zhihang Li$^\dagger$}
\author[1]{Yongbo Gai}
\author[2]{Zhengyi Yang}
\author[1]{Mengyu Zhou}
\author[1]{Erchao Zhao}
\author[1]{Xiaoxi Jiang}
\author[1]{Guanjun Jiang}
\affil[1]{Qwen Large Model Application Team, Alibaba}
\affil[2]{Institute of Automation, Chinese Academy of Sciences}
\affil[3]{Beijing University of Posts and Telecommunications}

\begin{abstract}
Vision-language process reward models (VL-PRMs) score intermediate reasoning steps and rerank candidates under test-time scaling, yet their step scores are often hard to interpret: a low reward may indicate a genuine reasoning mistake, or simply unreliable visual grounding by the policy or the verifier. This entanglement yields systematic false positives (rewarding hallucinated visual premises) and false negatives (penalizing correct grounded statements), degrading both reranking and error localization.

We propose \textbf{Explicit Visual Premise Verification (EVPV)}, a lightweight, test-time framework that decouples visual premise reliability from step correctness. EVPV prompts the policy to emit a step-wise visual checklist and independently extracts structured visual constraints from the image. By matching checklist claims against constraints, EVPV computes a scalar visual reliability signal and uses it to gate rewards for visually dependent steps, avoiding per-step tool calls. To fully realize this framework, we train \textbf{EVPV-PRM}, a Qwen2.5-VL-Instruct-7B-based step verifier that provides probabilistic base rewards and can be calibrated by EVPV at inference time. Across VisualProcessBench and six downstream benchmarks, EVPV improves step-level verification and yields overall Best-of-$N$ reranking gains, while our EVPV-PRM achieves strong performance as a deployable reranker. Under controlled constraint corruption, performance degrades monotonically, providing interventional evidence that the gains are driven by constraint fidelity.
\end{abstract}

\begin{document}
\maketitle

\section{Introduction}
\label{sec:introduction}

Multimodal mathematical reasoning couples two error-prone components:
\emph{visual perception} (e.g., diagrams, tables/OCR, geometric relations) and
\emph{symbolic reasoning} (derivation and computation). Although modern MLLMs
can generate fluent multi-step solutions, a single perceptual misread can derail
the whole chain while keeping later steps locally coherent, making
\emph{process-level} verification and selection crucial under test-time scaling
such as Best-of-$N$ and search-based decoding \citep{zheng2025survey,ma2023let,zhang2024rest}.

Process reward models (PRMs) implement process supervision by scoring each step,
and are widely used for Best-of-$N$ reranking, guided search, and post-training
\citep{zheng2025survey,ma2023let,zhang2024rest}. In vision-language reasoning,
VisualPRM and VisualProcessBench show that step-aware critics improve
performance under test-time scaling \citep{wang2025visualprm}, and data-efficient
training can further reduce verifier cost \citep{wang2025athena}. However,
current VL-PRMs often behave as \emph{black-box judges}: a low step score is
ambiguous---it may reflect a true reasoning error or unreliable visual grounding
by the policy/verifier---echoing broader concerns on PRM reliability and
calibration \citep{ye2025uncertainty,park2025know}.

\begin{figure*}[t]
  \centering
  \includegraphics[width=1\textwidth]{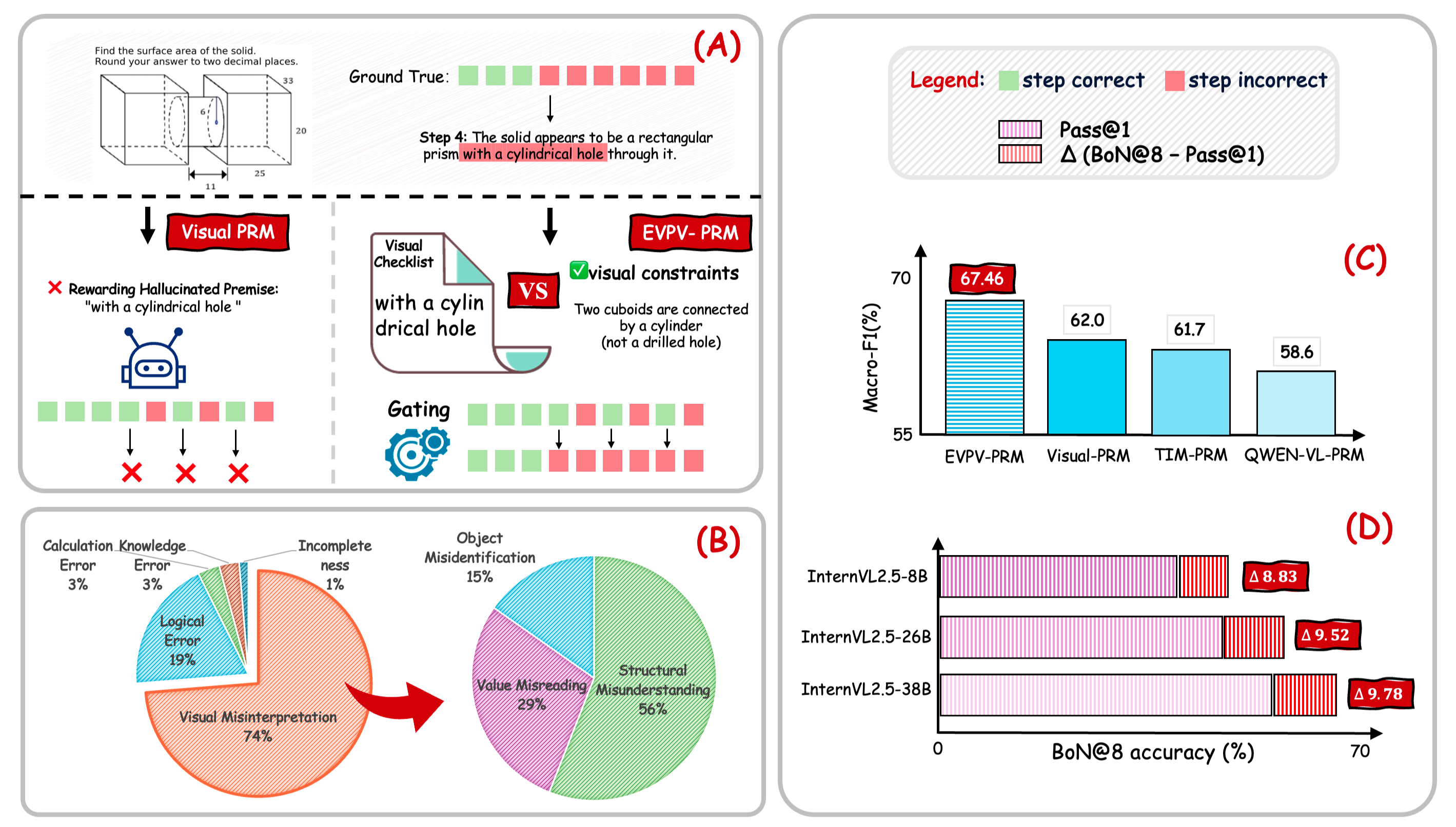}
  \caption{\textbf{EVPV} for reliable multimodal PRMs.
  \textbf{(A)} A motivating failure case (hallucinated visual premise).
  \textbf{(B)} Step errors are dominated by visual misinterpretation on VisualProcessBench.
  \textbf{(C)} EVPV-PRM improves step-level verification (Macro-F1).
  \textbf{(D)} EVPV-PRM improves Best-of-$8$ reranking (BoN@8) across InternVL2.5 scales.}
  \label{fig:teaser}
\end{figure*}

This ambiguity is a systematic source of verification errors. Under uncertain
grounding, a PRM may penalize correct visual statements (\emph{false negatives})
or reward hallucinated premises (\emph{false positives}), harming reranking and
error localization. Figure~\ref{fig:teaser}A shows such a case: VisualPRM rewards
a fluent step that assumes a nonexistent ``cylindrical hole.'' Visual
misinterpretation also constitutes a major fraction of step errors on
VisualProcessBench \citep{wang2025visualprm}. Tool-integrated verification can
reduce confirmation bias by querying evidence independently \citep{kuang2025tim},
but per-step tool calls are often too costly for long traces at Best-of-$N$
scale \citep{ma2023let,zhang2024rest}.

We propose \textbf{Explicit Visual Premise Verification (EVPV)} as a lightweight,
\emph{test-time} framework that decouples premise reliability from step
correctness. EVPV prompts the policy to state a step-wise \emph{visual checklist}
of explicit premises and uses a constraint extractor to predict \emph{structured}
visual facts (numeric readings, relations, and compositional structure) once per
instance. EVPV matches checklist claims against constraints to compute a visual
reliability signal, and uses it to gate rewards for visually dependent steps:
when premises are unreliable, rewards are attenuated toward neutrality; when
premises are reliable, base rewards are preserved.

To maximize the effectiveness of this framework in a deployable setting, we
train \textbf{EVPV-PRM}, a Qwen2.5-VL-Instruct-7B-based probabilistic step
verifier that provides base step rewards and can be calibrated by EVPV at
inference time. We evaluate EVPV on VisualProcessBench and six multimodal
reasoning benchmarks under Best-of-$N$ reranking. EVPV improves step-level
verification and yields overall reranking gains across InternVL2.5 policy
scales (Figure~\ref{fig:teaser}C--D). Finally, controlled corruption of the
extracted constraints induces monotonic performance degradation, providing
\emph{interventional} evidence that verification quality is driven by constraint
fidelity rather than incidental prompt effects.

\section{Related Work}
\label{sec:related_work}

\paragraph{Process reward models (PRMs).}
PRMs provide step-level supervision and are widely used for test-time scaling
(e.g., Best-of-$N$ reranking), guided decoding, and post-training
\citep{zheng2025survey,ma2023let,zhang2024rest}. Recent work explores stronger
verification procedures, including verifiers that generate intermediate analyses
or perform generative verification \citep{she2025r,zhao2025genprm,khalifa2025process,jia2025writing},
as well as improved learning objectives and supervision pipelines
\citep{yin2025dynamic,zhang2024entropy,zhang2025bidirectional,duan2025efficient,tan2025aurora,zhang2025openprm}.
Our work is orthogonal to these directions: rather than changing the verifier
architecture or training signal, we provide a \emph{test-time calibration
interface} that conditions step rewards on the reliability of the visual
premises they depend on.

\paragraph{Visual perception reliability and verification.}
MLLMs often struggle with fine-grained perception such as counting, geometry,
and structured reading \citep{fu2024blink,schulze2025visual}, motivating improved
vision encoders and perception-centric modeling/training
\citep{jain2024vcoder,yu2024texthawk,huang2023language,wu2024visionllm,huang2025visual,tang2024chain,yu2025introducing}.
These findings support our motivation: in multimodal reasoning, verification
should account for uncertainty in visual premises rather than treating all
image-conditioned statements as equally reliable.

\paragraph{Multimodal PRMs and grounded verification.}
VisualPRM introduces VisualPRM400K and VisualProcessBench, establishing a
standard training/evaluation pipeline for multimodal step verification
\citep{wang2025visualprm}. ATHENA improves data efficiency for training
multimodal PRMs \citep{wang2025athena}, and broader analyses study training
design choices and perception-focused supervision for VL-PRMs
\citep{ong2025training,luo2025unlocking,cao2025dreamprm}. EVPV builds on this
line by adding a \emph{grounding-aware calibration layer} on top of a step judge:
it makes visual premises explicit (a checklist), verifies them against
\emph{structured} visual constraints extracted once per instance, and gates
rewards accordingly. This is complementary to training better PRMs (e.g.,
VisualPRM/ATHENA): EVPV can be used with our trained EVPV-PRM or attached to an
external judge at inference time.

Tool-integrated verifiers such as TIM-PRM query visual evidence via tools to
reduce confirmation bias \citep{kuang2025tim}. EVPV targets a different
cost--reliability trade-off: it avoids \emph{per-step} tool calls by extracting
structured evidence once and reusing it across steps and candidates, making it
suitable for Best-of-$N$ reranking at scale. Recent works also strengthen
multimodal judges via richer reasoning, diagnosis, or correction (e.g., VRPRM
and GM-PRM) \citep{chen2025vrprm,zhang2025gm}; these approaches improve ``how the
judge reasons,'' whereas EVPV focuses on \emph{whether the visual premise is
trustworthy}, which is complementary when deep reasoning is performed on
unreliable premises. Finally, VLRMBench expands evaluation for vision-language
reward modeling beyond step verification \citep{ruan2025vlrmbench}; we primarily
use VisualProcessBench and downstream reranking benchmarks, and view broader
evaluation as future work.

\section{Methodology}
\label{sec:methodology}

\subsection{Problem Setup}
\label{sec:method:setup}
\begin{figure*}[t]
  \centering
  \includegraphics[width=\textwidth]{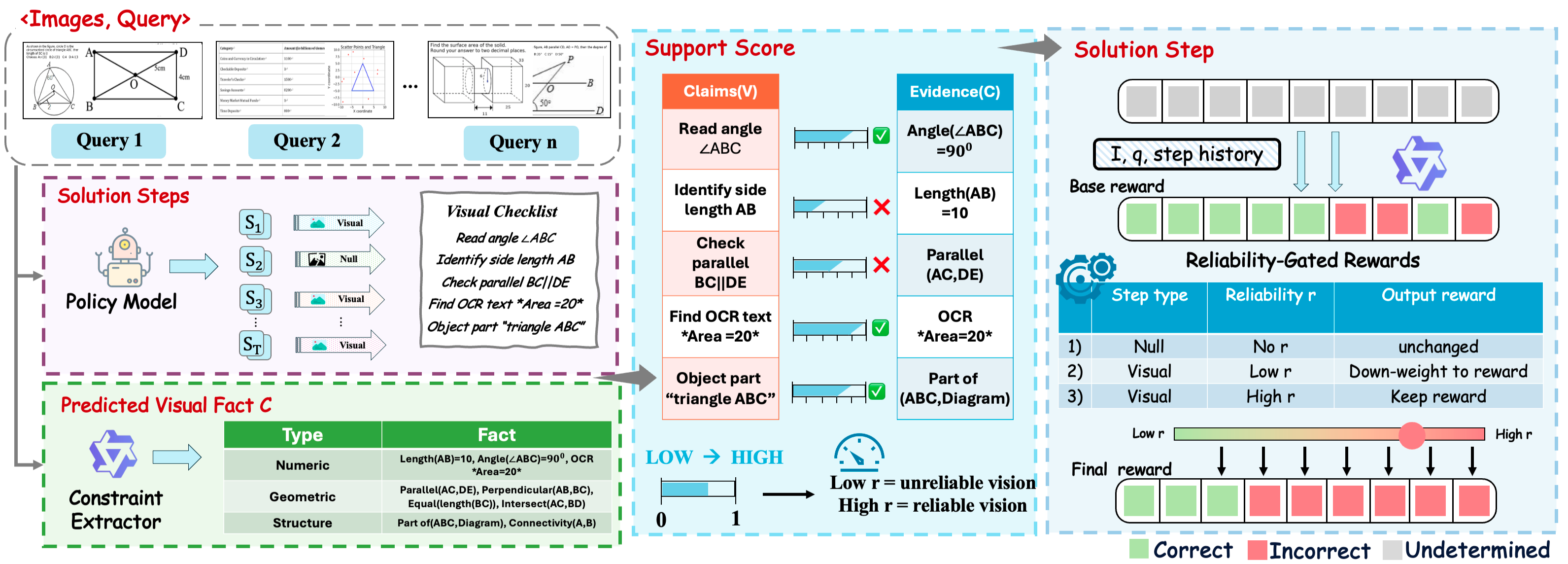}
  \caption{\textbf{Overview of EVPV-PRM.}
  Given an image $I$ and question $q$, the policy model generates a step-by-step solution and, for each step, declares whether it depends on visual evidence, forming a visual checklist of explicit claims.
  In parallel, a constraint extractor predicts a structured set of visual facts $C$ (numeric readings, geometric relations, and compositional structure).
  We compute a visual reliability score $r$ by matching checklist claims against $C$ to obtain support scores and aggregating them into a single confidence signal.
  A step verifier then produces base step rewards, which are calibrated by reliability gating: rewards for non-visual steps are kept unchanged, while rewards for visually dependent steps are down-weighted when $r$ is low and preserved when $r$ is high.
  The resulting reliability-gated step rewards are aggregated for Best-of-$N$ reranking and process diagnosis.}
  \label{fig:method_overview}
\end{figure*}

Each instance consists of an image $I$ and a question $q$.
A multimodal policy samples a step-by-step solution
$S=(s_1,\ldots,s_T)$ with a final answer $a$.
Our goal is \emph{premise-aware process verification}: assign a reward
$R_t\in[-1,1]$ to each step $s_t$ such that (i) step-level judgments are robust
to visual misperception, and (ii) aggregated trajectory scores support reliable
Best-of-$N$ reranking.

A key challenge in multimodal reasoning is that failures arise from two
distinct sources: \emph{visual grounding errors} (misread values, incorrect
relations/structure) and \emph{symbolic reasoning errors} (invalid deductions,
arithmetic mistakes). Standard VL-PRMs entangle the two by directly scoring
steps as if the underlying visual premises were reliable. We instead explicitly
verify the visual premises a trace relies on and use this signal to calibrate
step rewards at inference time.

\subsection{EVPV: Explicit Visual Premise Verification}
\label{sec:method:evpv}

EVPV is a lightweight, \emph{judge-agnostic} calibration interface for multimodal
process reward modeling. It operates in three stages:
(1) make each step's visual premises explicit (a \emph{checklist}),
(2) extract \emph{structured visual evidence} once per instance,
and (3) convert checklist--evidence consistency into a scalar \emph{visual
reliability} used to calibrate step rewards. Figure~\ref{fig:method_overview}
illustrates the pipeline.

\subsubsection{Step-wise Visual Checklist}
\label{sec:method:checklist}

We prompt the policy to accompany each reasoning step $s_t$ with a minimal
declaration of the \emph{visual premise} it depends on:
\begin{equation}
d_t \in \{\text{a verifiable visual assertion},\ \texttt{null}\}.
\end{equation}
If $d_t\neq \texttt{null}$, the step asserts dependence on a concrete visual
fact (e.g., ``the radius is $2$'', ``$AB\perp CD$'', ``a cone is attached on top
of a cylinder''). We define the visual-dependency indicator:
\begin{equation}
\nu_t=\mathbb{I}[d_t\neq\texttt{null}] \in \{0,1\}.
\end{equation}
Collecting all non-null declarations yields a \emph{visual checklist}
$V=\{v_j\}_{j=1}^{M}$. This checklist is the interface EVPV needs: it turns
implicit visual assumptions into explicit claims that can be verified
independently from later algebra. We empirically audit the completeness of this
policy-reported visual dependency signal (i.e., omission vs.\ over-reporting) on
a human-verified subset; see Appendix~\ref{app:checklist_completeness} and
Table~\ref{tab:checklist_audit}.

\subsubsection{Structured Visual Evidence (Constraints)}
\label{sec:method:constraints}

To verify the checklist, we extract structured visual evidence once per
instance using a constraint extractor $E_\phi$:
\begin{equation}
C = E_\phi(I,q) = \{c_k\}_{k=1}^{K}.
\end{equation}
Each constraint follows a unified JSON schema (Appendix~A) covering
(i) numeric readings (lengths, angles, table entries),
(ii) relations (parallel/perpendicular/equality/incidence/containment), and
(iii) compositional structure (part--whole, attachments, adjacency).
At test time, EVPV relies only on predicted $C$; no gold facts are used.
Crucially, $C$ is computed \emph{once} and reused across all steps and all
candidates for the same $(I,q)$, enabling scalable Best-of-$N$ reranking. We
evaluate the fidelity of these predicted constraints with a human-annotated
study, reporting precision/recall/F1 by category; see
Appendix~\ref{app:extractor_fidelity} and
Table~\ref{tab:extractor_fidelity_prf}.

\subsubsection{Consistency-to-Reliability}
\label{sec:method:reliability}

EVPV converts checklist--evidence consistency into a scalar \emph{visual
reliability} score. Let $m(\cdot)$ be a type-aware matching function that
measures whether a checklist claim is supported by the extracted constraints
$C$:
\begin{equation}
p_j = m(v_j, C)\in[0,1],
\end{equation}
where $p_j$ is high when the claim is entailed by $C$ (with numeric tolerance
and entity/relation alignment; Appendix~B).

We then aggregate per-claim support scores $\{p_j\}_{j=1}^{M}$ into a single
candidate-level reliability score $r\in[0,1]$ using a smoothed geometric mean:
\begin{equation}
r \;=\; \exp\!\left(\frac{1}{M}\sum_{j=1}^{M}\log(\epsilon+p_j)\right),
\label{eq:geomagg}
\end{equation}
where $\epsilon$ is a small constant for numerical stability. This aggregation
is intentionally sensitive to catastrophic premise failures: if any claim is
strongly unsupported ($p_j\approx 0$), $r$ drops sharply, reflecting that a
single misread visual premise can invalidate the entire reasoning trace.

\subsection{Judge-Agnostic Step Rewards with Reliability Gating}
\label{sec:method:gating}

\subsubsection{A unified judge interface}
\label{sec:method:judge_interface}

EVPV does not assume a particular step judge. Instead, it takes as input a
\emph{base step reward} from an arbitrary judge module $J$:
\begin{equation}
R_t^{\mathrm{base}} = J(I,q,s_{\le t}) \in [-1,1].
\label{eq:judge_interface}
\end{equation}
This abstraction lets EVPV act as a plug-in calibration layer on top of both
trained PRMs and prompted black-box judges.

\paragraph{Instantiation 1: EVPV-PRM (our trained verifier).}
In our main system, the judge is a trained step verifier $V_\theta$ that outputs
a correctness probability:
\begin{equation}
u_t = P_{\theta}(y_t=1\mid I,q,s_{\le t})\in[0,1],
\end{equation}
mapped to a signed reward:
\begin{equation}
R_t^{\mathrm{base}} = 2u_t-1.
\label{eq:base_reward}
\end{equation}

\paragraph{Instantiation 2: EVPV as a plug-in for external VLM/LLM judges.}
EVPV can also be applied to an external prompted judge (e.g., GPT/Gemini) that
outputs a binary step decision, yielding
$R_t^{\mathrm{base}}\in\{-1,+1\}$.
All subsequent EVPV computations (reliability, gating, aggregation) remain
unchanged.

\subsubsection{Reliability gating}
\label{sec:method:reliability_gating}

A base step reward alone is ambiguous in multimodal settings: a low score may
reflect a true reasoning error, or simply that the step rests on an unreliable
visual premise. EVPV resolves this ambiguity by \emph{calibrating} rewards for
visually dependent steps using $r$.

We convert reliability into a smooth gating factor:
\begin{equation}
\alpha(r)=\sigma\!\big(\beta(r-\tau)\big)\in(0,1),
\end{equation}
where $\tau$ is a reliability threshold, $\beta$ controls sharpness, and
$\sigma$ is the logistic function. The final step reward is
\begin{equation}
R_t=
\begin{cases}
R_t^{\mathrm{base}}, & \nu_t=0,\\
\alpha(r)\,R_t^{\mathrm{base}}, & \nu_t=1.
\end{cases}
\label{eq:gated_reward}
\end{equation}
Intuitively, when $r$ is low, $\alpha(r)\approx 0$ and visually grounded steps
are pushed toward a neutral score, preventing unreliable visual premises from
producing overly confident positive/negative signals. When $r$ is high,
$\alpha(r)\approx 1$ and the judge behaves like a conventional PRM.

\subsection{Trajectory Scoring for Best-of-$N$ Reranking}
\label{sec:method:traj_score}

Given a candidate solution $S$, we compute gated step rewards
$\{R_t\}_{t=1}^{T}$ and aggregate them into a trajectory score for reranking.
Since reliability gating rescales reward magnitudes, we use a
\emph{magnitude-sensitive} aggregation so that gating can influence candidate
ranking.
We first map $R_t\in[-1,1]$ to positive values:
\begin{equation}
\tilde{R}_t = \epsilon + \frac{R_t+1}{2}\in(\epsilon,1+\epsilon),
\end{equation}
and compute a geometric-mean trajectory score:
\begin{equation}
\mathrm{Score}(S)
=\exp\!\left(\frac{1}{T}\sum_{t=1}^{T}\log \tilde{R}_t\right).
\label{eq:traj_score}
\end{equation}
We select the candidate with the highest $\mathrm{Score}(S)$.
Alternative aggregations are reported in Appendix~E.

\begin{figure}[!b]
    \centering
    \includegraphics[width=\linewidth]{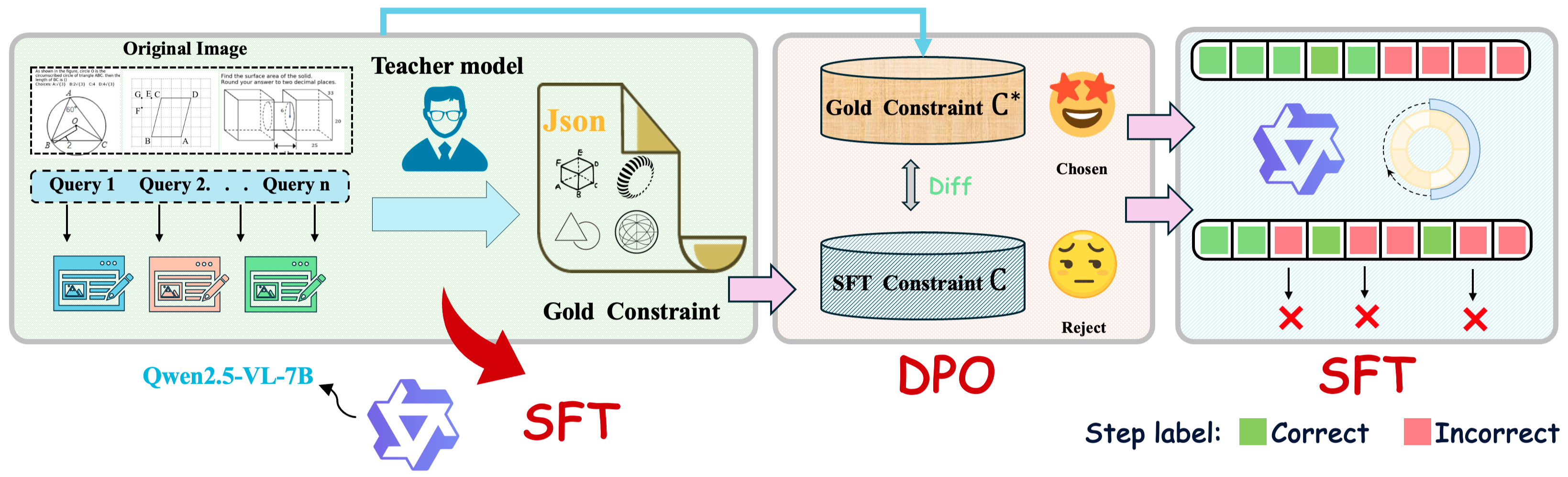}
    \caption{Training and inference pipeline of EVPV. EVPV trains a constraint extractor $E_\phi$ and a step verifier $V_\theta$. The policy is not trained and is only prompted to produce step-wise solutions with a visual checklist at inference time.}
    \label{fig:training_pipeline}
\end{figure}

\subsection{Training and Inference}
\label{sec:method:training}

EVPV introduces two trainable modules: the constraint extractor $E_\phi$ and the
step verifier $V_\theta$ (used in the EVPV-PRM instantiation). The policy is
not trained in this work; it is only prompted to output steps and checklist
items at inference time (Figure~\ref{fig:training_pipeline}).

\paragraph{Training data and evaluation boundary.}
To avoid any training--test contamination, we \textbf{do not use VisualProcessBench}
\citep{wang2025visualprm} for training.
All training data are sampled from \textbf{VisualPRM400K} \citep{wang2025visualprm},
restricted to geometry- and table-centric subsets
(\textsc{Geo170K}, \textsc{GeometryData}, \textsc{GeomVerse}, \textsc{GEOS},
\textsc{MAVIS-Geometry}, \textsc{TabMWP}, \textsc{UniGeo}).
VisualProcessBench is used \textbf{only} for step-level evaluation
(Section~\ref{sec:exp:vpbench}; Table~\ref{tab:vpbench}).

\paragraph{Training the constraint extractor.}
We distill pseudo-gold structured constraints $C^\star$ using a strong teacher
Qwen3-vl-235b-a22b-instruct. Specifically, we construct an SFT corpus
of 26{,}454 image--question pairs from the above VisualPRM400K subsets; for each
pair, the teacher is prompted with our JSON schema (Appendix~A) and the
\emph{correct solution steps} to produce a solution-critical constraint set
$C^\star$. We fine-tune $E_\phi$ by maximizing $P_\phi(C^\star\mid I,q)$:
\begin{equation}
\mathcal{L}_{\mathrm{con}}(\phi)=-\log P_{\phi}(C^\star\mid I,q).
\label{eq:con_sft}
\end{equation}
To further improve fidelity, we run a DPO stage on an additional 5{,}832
instances where the teacher rewrites constraints to form preference pairs
(Appendix~\ref{app:training}).

\definecolor{headerblue}{HTML}{DCEAF7}      
\definecolor{groupblue}{HTML}{EAF2FB}       
\definecolor{evpvgreen}{HTML}{EAF6EA}       
\definecolor{deltagray}{HTML}{F7F7F7}       
\definecolor{finalgreen}{HTML}{DFF0DF}      
\definecolor{posgain}{HTML}{1E88E5}         
\definecolor{neggain}{HTML}{C62828}         
\definecolor{textgray}{HTML}{4D4D4D}        

\sisetup{
  detect-weight=true,
  detect-family=true,
  table-number-alignment = center,
  table-format = 2.2
}

\newcommand{\posdelta}[1]{\textcolor{posgain}{\textbf{#1}}}
\newcommand{\negdelta}[1]{\textcolor{neggain}{#1}}
\newcommand{\cmark}{\ding{51}}
\newcommand{\xmark}{\ding{55}}

\begin{table*}[t]
\centering
\caption{\textbf{VisualProcessBench Macro-F1 (\%).}
For external judge models, \xmark\ uses the judge alone; \cmark\ attaches the EVPV plug-in (constraints-based reliability + gating, with the same constraints provided as evidence).
$\Delta$ = \cmark\ $-$ \xmark\ (points); positive $\Delta$ is shown in blue and negative $\Delta$ in red.}
\label{tab:vpbench}
\small
\setlength{\tabcolsep}{5.2pt}
\resizebox{\textwidth}{!}{%
\begin{tabular}{@{}ll
S[table-format=2.2]
S[table-format=2.2]
S[table-format=2.2]
S[table-format=2.2]
S[table-format=2.2]
S[table-format=2.2]@{}}
\toprule
\rowcolor{headerblue}
\textbf{Model} & \textbf{EVPV} &
\textbf{DynaMath} & \textbf{MMMU} & \textbf{MathVerse} & \textbf{MathVision} & \textbf{WeMath} & \textbf{Overall} \\
\midrule

\rowcolor{groupblue}
\multicolumn{8}{c}{\textbf{Proprietary Models}} \\
\midrule

gpt-4o-mini
 & \xmark  & 56.57 & 54.08 & 52.53 & 51.42 & 56.74 & 53.57 \\
\rowcolor{evpvgreen}
 & \cmark  & 58.13 & 53.20 & 54.09 & 52.07 & 54.62 & 54.29 \\
\rowcolor{deltagray}
 & $\Delta$ &
 \posdelta{+1.56} & \negdelta{-0.88} & \posdelta{+1.56} & \posdelta{+0.65} & \negdelta{-2.12} & \posdelta{+0.72} \\
\addlinespace[2pt]

doubao-seed-1.6-vision
 & \xmark  & 66.19 & 59.47 & 63.12 & 61.07 & 62.74 & 62.77 \\
\rowcolor{evpvgreen}
 & \cmark  & 68.66 & 61.86 & 65.57 & 62.51 & 64.62 & 64.91 \\
\rowcolor{deltagray}
 & $\Delta$ &
 \posdelta{+2.47} & \posdelta{+2.39} & \posdelta{+2.45} & \posdelta{+1.44} & \posdelta{+1.88} & \posdelta{+2.14} \\
\addlinespace[2pt]

Gemini 2.5 Pro
 & \xmark  & 68.47 & 63.34 & 68.26 & 65.15 & 69.48 & 67.13 \\
\rowcolor{evpvgreen}
 & \cmark  & 71.32 & 64.42 & 69.78 & 65.26 & 72.43 & 68.64 \\
\rowcolor{deltagray}
 & $\Delta$ &
 \posdelta{+2.85} & \posdelta{+1.08} & \posdelta{+1.52} & \posdelta{+0.11} & \posdelta{+2.95} & \posdelta{+1.51} \\
\midrule

\rowcolor{groupblue}
\multicolumn{8}{c}{\textbf{Open-source Models}} \\
\midrule

qwen2.5-vl-72b-instruct
 & \xmark  & 56.99 & 59.43 & 56.43 & 58.09 & 55.72 & 57.19 \\
\rowcolor{evpvgreen}
 & \cmark  & 61.43 & 60.25 & 59.85 & 59.12 & 59.72 & 59.99 \\
\rowcolor{deltagray}
 & $\Delta$ &
 \posdelta{+4.44} & \posdelta{+0.82} & \posdelta{+3.42} & \posdelta{+1.03} & \posdelta{+4.00} & \posdelta{+2.80} \\
\addlinespace[2pt]

Qwen3-VL-30B-A3B-instruct
 & \xmark  & 58.95 & 61.29 & 57.37 & 57.49 & 58.76 & 58.22 \\
\rowcolor{evpvgreen}
 & \cmark  & 62.27 & 59.00 & 59.68 & 56.49 & 59.50 & 59.26 \\
\rowcolor{deltagray}
 & $\Delta$ &
 \posdelta{+3.32} & \negdelta{-2.29} & \posdelta{+2.31} & \negdelta{-1.00} & \posdelta{+0.74} & \posdelta{+1.04} \\
\addlinespace[2pt]

Qwen3-VL-235B-A22B-instruct
 & \xmark  & 57.63 & 58.73 & 58.08 & 59.59 & 58.76 & 58.51 \\
\rowcolor{evpvgreen}
 & \cmark  & 68.43 & 61.54 & 65.97 & 64.54 & 64.16 & 65.45 \\
\rowcolor{deltagray}
 & $\Delta$ &
 \posdelta{+10.80} & \posdelta{+2.81} & \posdelta{+7.89} & \posdelta{+4.95} & \posdelta{+5.40} & \posdelta{+6.94} \\
\midrule

\rowcolor{groupblue}
\multicolumn{8}{c}{\textbf{Process Reward Models}} \\
\midrule

\multicolumn{2}{@{}l}{QWEN-VL-PRM-7B \citep{ong2025training}} & 58.30 & 55.80 & 58.80 & 55.70 & 59.80 & 58.60 \\
\multicolumn{2}{@{}l}{TIM-PRM-8B \citep{kuang2025tim}}        & 65.90 & 58.30 & 61.90 & 58.30 & 63.90 & 61.70 \\
\multicolumn{2}{@{}l}{VisualPRM-8B \citep{wang2025visualprm}} & 62.70 & 58.50 & 61.00 & 62.10 & 61.80 & 62.00 \\
\rowcolor{finalgreen}
\multicolumn{2}{@{}l}{\textbf{EVPV-PRM}}                     &
\bfseries 69.57 & \bfseries 68.86 & \bfseries 67.09 & \bfseries 65.27 & \bfseries 69.11 & \bfseries 67.46 \\
\bottomrule
\end{tabular}%
}
\end{table*}

\paragraph{Training the step verifier (EVPV-PRM).}
We train $V_\theta$ as a probabilistic step verifier using 19{,}490 supervised
step-labeled trajectories sampled from VisualPRM400K.
Each training trajectory provides step-level correctness labels derived from its
reference solution trace, yielding binary targets $y_t\in\{0,1\}$.
We optimize binary cross-entropy:
\begin{equation}
\mathcal{L}_{V}(\theta)=
-\sum_{t=1}^{T}\Big(
y_t\log u_t+(1-y_t)\log(1-u_t)
\Big),
\label{eq:verifier_bce}
\end{equation}
where $u_t=P_\theta(y_t=1\mid I,q,s_{\le t})$.
\emph{Reliability gating is applied only at inference time} (Eq.~\ref{eq:gated_reward}),
keeping verifier training unchanged and allowing EVPV to be attached to other
judges.

\paragraph{Inference.}
For each $(I,q)$, we first predict constraints $C=E_\phi(I,q)$ once. For each
candidate solution $S$, we (i) parse the policy-produced checklist $\{d_t\}$ to
obtain $\{\nu_t\}$ and $V$, (ii) compute reliability $r$ by matching $V$ against
$C$, (iii) obtain base step rewards $R_t^{\mathrm{base}}$ from a judge $J$ (our
trained $V_\theta$ or an external prompted judge), (iv) apply reliability gating
to produce $\{R_t\}$, and (v) aggregate $\{R_t\}$ using
Eq.~\ref{eq:traj_score} for reranking. This yields premise-aware verification
without per-step tool calls.
\section{Experiments}
\label{sec:experiments}

\subsection{Benchmarks, Protocol, and Baselines}
\label{sec:exp:setup}

We evaluate EVPV from two angles: (i) \emph{step-level verification} on annotated reasoning traces, and (ii) \emph{deployable test-time gains} under Best-of-$N$ reranking.
For step-level evaluation we use \textbf{VisualProcessBench} \citep{wang2025visualprm}.
For downstream evaluation we use six multimodal reasoning benchmarks: LogicVista \citep{xiao2024logicvista}, MMMU \citep{yue2024mmmu}, MathVerse-VO \citep{zhang2024mathverse}, MathVision \citep{wang2024measuring}, MathVista \citep{lu2023mathvista}, and WeMath \citep{qiao2025we}.

\paragraph{Evaluation protocol (what is fixed vs.\ what changes).}
All experiments follow the EVPV pipeline in Section~\ref{sec:methodology}:
(i) a policy produces step traces with per-step \texttt{visualdependency} (our visual checklist),
(ii) a constraint extractor predicts structured constraints $C$ once per instance,
(iii) checklist--constraint matching yields a candidate-level reliability $r$,
and (iv) reliability gating calibrates base step rewards for visually dependent steps.
What differs across settings is \emph{how the base step reward is produced}:

\begin{itemize}
  \item \textbf{EVPV-PRM (ours).} Base rewards $R_t^{\mathrm{base}}\in[-1,1]$ are produced by our trained probabilistic step verifier $V_\theta$ (Eq.~\ref{eq:base_reward}), then gated by EVPV.
  \item \textbf{EVPV plug-in for external judges.} For a black-box VLM/LLM judge (e.g., GPT/Gemini/Qwen) that outputs a binary step judgment
  $R_t^{\mathrm{base}}\in\{-1,+1\}$, we can attach EVPV as a \emph{plug-in calibration layer}. In Table~\ref{tab:vpbench}, \texttt{No} means using the external judge \emph{alone}, while \texttt{Yes} means using the same external judge \emph{with EVPV enabled} (i.e., computing $r$ from structured constraints and applying reliability gating to the judge's step rewards).%
  \footnote{In the \texttt{Yes} setting, the judge additionally receives our predicted structured constraints as evidence (Appendix~D), and EVPV uses the same constraints to compute $r$; this keeps the plug-in setting self-contained and fully inference-time.}
\end{itemize}

\paragraph{Metrics.}
On VisualProcessBench we report step-level \textbf{Macro-F1} (primary) and accuracy.
On downstream benchmarks we report \textbf{Pass@1} (policy accuracy without reranking), \textbf{BoN@k} (accuracy after reranking $k$ samples), and the practical gain $\Delta_k=\mathrm{BoN@k}-\mathrm{Pass@1}$.
We also report \textbf{Std Pass@k}, the oracle upper bound of the candidate set, to separate candidate quality from selection quality.

\definecolor{headerblue}{HTML}{DCEAF7}     
\definecolor{groupblue}{HTML}{EEF3F8}      
\definecolor{visualprmrow}{HTML}{F6F1E8}   
\definecolor{evpvrow}{HTML}{E8F4EA}        
\definecolor{deltarow}{HTML}{F8F8F8}       
\definecolor{posgain}{HTML}{1E88E5}        
\definecolor{modelgray}{HTML}{666666}      

\newcommand{\imp}[1]{\textcolor{posgain}{\textbf{#1}}}

\begin{table*}[t]
\centering
\caption{\textbf{Downstream Best-of-8 reranking with InternVL2.5 policies.}
BoN@8 accuracy (\%) after reranking with different PRMs; numbers denote $\Delta_8$ (BoN@8 $-$ Pass@1) for our PRM.}
\label{tab:bon}
\small
\setlength{\tabcolsep}{5pt}
\renewcommand{\arraystretch}{1.12}
\resizebox{\textwidth}{!}{%
\begin{tabular}{lccccccc}
\toprule
\rowcolor{headerblue}
\textbf{Model} & \textbf{MathVista} & \textbf{MathVision} & \textbf{MathVerse-VO} & \textbf{WeMath} & \textbf{LogicVista} & \textbf{MMMU} & \textbf{Overall} \\
\midrule

\rowcolor{groupblue}
\multicolumn{8}{c}{\textbf{Proprietary Models}} \\
\midrule
GPT-4o            & 60.00 & 31.20 & 40.60 & 45.80 & 52.80 & 70.70 & 47.90 \\
Gemini-2.0-Flash  & 70.40 & 43.60 & 47.80 & 47.40 & 52.30 & 69.90 & 53.40 \\
Claude-3.5-Sonnet & 65.30 & 35.60 & 46.30 & 44.00 & 60.40 & 66.40 & 50.50 \\
\midrule

\rowcolor{groupblue}
\multicolumn{8}{c}{\textbf{Open-source Models}} \\
\midrule

InternVL2.5-8B     & 64.50 & 17.00 & 22.80 & 23.50 & 36.38 & 56.20 & 32.84 \\
\rowcolor{visualprmrow}
+VisualPRM         & 68.50 & 25.70 & 35.80 & 36.50 & 43.80 & 60.20 & 41.40 \\
\rowcolor{deltarow}
                   & +4.00 & +8.70 & +13.00 & +13.00 & +7.80 & +4.00 & +8.40 \\
\rowcolor{evpvrow}
\textbf{+EVPV-PRM} & \textbf{76.30} & \textbf{22.07} & \textbf{29.47} & \textbf{37.45} & \textbf{45.33} & \textbf{67.75} & \textbf{41.67} \\
\rowcolor{deltarow}
                   & \imp{+11.80} & \imp{+5.07} & \imp{+6.67} & \imp{+13.95} & \imp{+8.95} & \imp{+11.55} & \imp{+8.83} \\
\midrule

InternVL2.5-26B    & 68.20 & 23.40 & 24.00 & 30.90 & 39.64 & 60.70 & 37.23 \\
\rowcolor{visualprmrow}
+VisualPRM         & 73.10 & 29.60 & 39.10 & 40.80 & 51.00 & 63.90 & 45.80 \\
\rowcolor{deltarow}
                   & +4.90 & +6.20 & +15.10 & +9.90 & +11.40 & +3.20 & +8.90 \\
\rowcolor{evpvrow}
\textbf{+EVPV-PRM} & \textbf{79.60} & \textbf{28.11} & \textbf{32.47} & \textbf{42.14} & \textbf{51.72} & \textbf{69.25} & \textbf{46.75} \\
\rowcolor{deltarow}
                   & \imp{+11.40} & \imp{+4.71} & \imp{+8.47} & \imp{+11.24} & \imp{+12.08} & \imp{+8.55} & \imp{+9.52} \\
\midrule

InternVL2.5-38B    & 71.90 & 32.20 & 36.90 & 38.30 & 47.90 & 63.90 & 45.44 \\
\rowcolor{visualprmrow}
+VisualPRM         & 73.90 & 35.20 & 46.70 & 46.20 & 53.70 & 69.00 & 50.70 \\
\rowcolor{deltarow}
                   & +2.00 & +3.00 & +9.80 & +7.90 & +5.80 & +5.10 & +6.30 \\
\rowcolor{evpvrow}
\textbf{+EVPV-PRM} & \textbf{83.50} & \textbf{37.59} & \textbf{47.67} & \textbf{50.00} & \textbf{58.74} & \textbf{72.33} & \textbf{55.22} \\
\rowcolor{deltarow}
                   & \imp{+11.60} & \imp{+5.39} & \imp{+10.77} & \imp{+11.70} & \imp{+10.84} & \imp{+8.43} & \imp{+9.78} \\
\bottomrule
\end{tabular}%
}
\end{table*}

\paragraph{Trajectory scoring for reranking.}
Unless stated otherwise, we use the geometric-mean aggregation in Eq.~\ref{eq:traj_score} for Best-of-$N$ reranking (Appendix~E reports alternatives). This magnitude-sensitive aggregation ensures that reliability gating can affect candidate ranking.

\paragraph{Baselines.}
We compare against multimodal PRMs including \textbf{VisualPRM} \citep{wang2025visualprm}, \textbf{QWEN-VL-PRM-7B} \citep{ong2025training} and the tool-integrated verifier \textbf{TIM-PRM} \citep{kuang2025tim}.
We also evaluate several strong MLLMs as step judges under a standardized prompt, with two conditions: \texttt{No} (judge alone) and \texttt{Yes} (judge + EVPV plug-in).
Finally, we include component ablations of EVPV (checklist, constraints, matching, gating).
We also report an efficiency comparison with tool-integrated verification:
Table~\ref{tab:cost} (Appendix G) summarizes the per-question inference cost in terms of model/tool calls and a unified token/latency accounting for EVPV and TIM-PRM.

\subsection{Exp-1: Step Verification on VisualProcessBench}
\label{sec:exp:vpbench}

We evaluate step-level verification directly on VisualProcessBench \citep{wang2025visualprm}.
Table~\ref{tab:vpbench} compares our method with prior multimodal PRMs and a set of judge models.

\paragraph{Protocol.}
For each annotated trace, we evaluate each step $s_t$ given $(I,q,s_{\le t})$ and obtain a binary correctness prediction.
For PRM-style verifiers (including ours), we threshold the predicted probability at $0.5$ (equivalently, $R_t^{\mathrm{base}}>0$).
For external judge models, we use their prompted binary output in $\{+1,-1\}$.
Macro-F1 is computed over all steps and then macro-averaged across subsets as in \citet{wang2025visualprm}.

\paragraph{Judge models: \texttt{No} vs.\ \texttt{Yes}.}
For judge models in Table~\ref{tab:vpbench}, \texttt{No} uses the external judge directly to label each step.
\texttt{Yes} attaches EVPV as a plug-in: for each instance we predict structured constraints $C=E_\phi(I,q)$ once, compute reliability $r$ by matching the policy-produced checklist against $C$ (Section~\ref{sec:method:reliability}), and apply reliability gating to calibrate the judge's step rewards before thresholding. This evaluates whether premise-aware calibration improves step discrimination for black-box judges under real visual uncertainty.

Two observations stand out in Table~\ref{tab:vpbench}.
First, our method achieves the best overall Macro-F1 among the compared PRMs, indicating stronger step discrimination under real visual uncertainty.
Second, many judge models improve under \texttt{Yes}, suggesting that EVPV-style premise verification and reliability calibration is broadly reusable as an inference-time plug-in---even without retraining the judge---and that a non-trivial part of verification error comes from missing or unreliable grounding.

\definecolor{headerblue}{HTML}{DCEAF7}     
\definecolor{groupblue}{HTML}{EEF3F8}      
\definecolor{fullgreen}{HTML}{E8F4EA}      
\definecolor{overallgray}{HTML}{F5F5F5}    
\definecolor{posgain}{HTML}{1E88E5}        
\definecolor{neggain}{HTML}{C62828}        
\definecolor{bestgreen}{HTML}{2E7D32}      

\sisetup{
  detect-weight=true,
  detect-family=true,
  table-number-alignment=center
}

\newcommand{\abpos}[1]{\textcolor{posgain}{\textbf{#1}}}
\newcommand{\abneg}[1]{\textcolor{neggain}{#1}}

\begin{table*}[t]
\centering
\caption{\textbf{Key ablations on VisualProcessBench} (Macro-F1; higher is better). $\Delta$ is relative to the full method.}
\label{tab:ablation_vpbench_key}
\small
\setlength{\tabcolsep}{5pt}
\renewcommand{\arraystretch}{1.12}
\resizebox{\textwidth}{!}{%
\begin{tabular}{@{}l
S[table-format=2.2]
S[table-format=2.2]
S[table-format=2.2]
S[table-format=2.2]
S[table-format=2.2]
S[table-format=2.2]
S[table-format=+2.2]
@{}}
\toprule
\rowcolor{headerblue}
\textbf{Variant} &
\textbf{DynaMath} & \textbf{MMMU} & \textbf{MathVerse} &
\textbf{MathVision} & \textbf{WeMath} & \textbf{Overall} &
\textbf{$\Delta$} \\
\midrule

\rowcolor{groupblue}
\multicolumn{8}{@{}l}{\textbf{Full Method}} \\
\rowcolor{fullgreen}
\textbf{Full (EVPV + gating)} &
\bfseries 69.57 & \bfseries 68.86 & \bfseries 67.09 &
\bfseries 65.27 & \bfseries 69.11 & \bfseries 67.46 &
\bfseries \abpos{+0.00} \\
\addlinespace[2pt]

\rowcolor{groupblue}
\multicolumn{8}{@{}l}{\textbf{Evidence / structure ablations}} \\
w/o structured facts (caption-only)     & 67.75 & 58.09 & 63.48 & 60.68 & 67.10 & 63.38 & \abneg{-4.08} \\
w/o constraints (facts = $\varnothing$) & 66.66 & 55.80 & 62.61 & 59.13 & 65.81 & 62.11 & \abneg{-5.35} \\
w/ shuffled facts (structure corrupted) & 62.86 & 52.57 & 59.81 & 58.52 & 64.77 & 59.82 & \abneg{-7.64} \\
\addlinespace[2pt]

\rowcolor{groupblue}
\multicolumn{8}{@{}l}{\textbf{Remove modalities / severe corruption}} \\
w/o vision (text-only judge, keep JSON) & 58.44 & 49.44 & 53.59 & 54.07 & 61.02 & 54.93 & \abneg{-12.53} \\
w/o vision \& w/o JSON (text-only)      & 54.49 & 43.93 & 42.78 & 50.84 & 53.78 & 48.23 & \abneg{-19.23} \\
w/ drop-facts corruption                & 34.90 & 34.40 & 36.29 & 36.14 & 35.96 & 35.77 & \abneg{-31.69} \\
\bottomrule
\end{tabular}%
}
\end{table*}

\subsection{Exp-2: Best-of-$N$ Reranking in Downstream Benchmarks}
\label{sec:exp:bon}

We next test whether premise-aware verification translates into deployable test-time gains.
We rerank candidates generated by InternVL2.5 policy models at three scales (8B/26B/38B).
For each question, the policy samples $k\in\{1,\ldots,8\}$ candidate solutions; we rerank them using step rewards and report BoN@8.

\paragraph{Protocol.}
For each question, we first predict structured constraints $C=E_\phi(I,q)$ once.
For each candidate solution, the policy provides steps with per-step \texttt{visualdependency}.
We compute candidate-level reliability $r$ via checklist--constraint matching, obtain base step rewards from the reranker PRM (VisualPRM or our EVPV-PRM), apply EVPV gating when applicable, and aggregate gated rewards using the geometric-mean trajectory score (Eq.~\ref{eq:traj_score}) to rank the $8$ candidates.

Table~\ref{tab:bon} summarizes the results.
Across all three policy sizes, our PRM yields consistent gains over the base policy and improves upon VisualPRM \citep{wang2025visualprm} in overall performance (e.g., +8.83, +9.52, and +9.78 points over Pass@1 for 8B/26B/38B, respectively).
The improvements are especially pronounced on visually intensive benchmarks such as MathVista, WeMath, and LogicVista, which matches EVPV’s intent: when early visual premises are the dominant failure mode, reliability-aware step scoring reduces selection errors without incurring the per-step tool overhead of TIM-PRM \citep{kuang2025tim}.

\begin{figure}[t]
  \centering
  \includegraphics[width=0.8\linewidth]{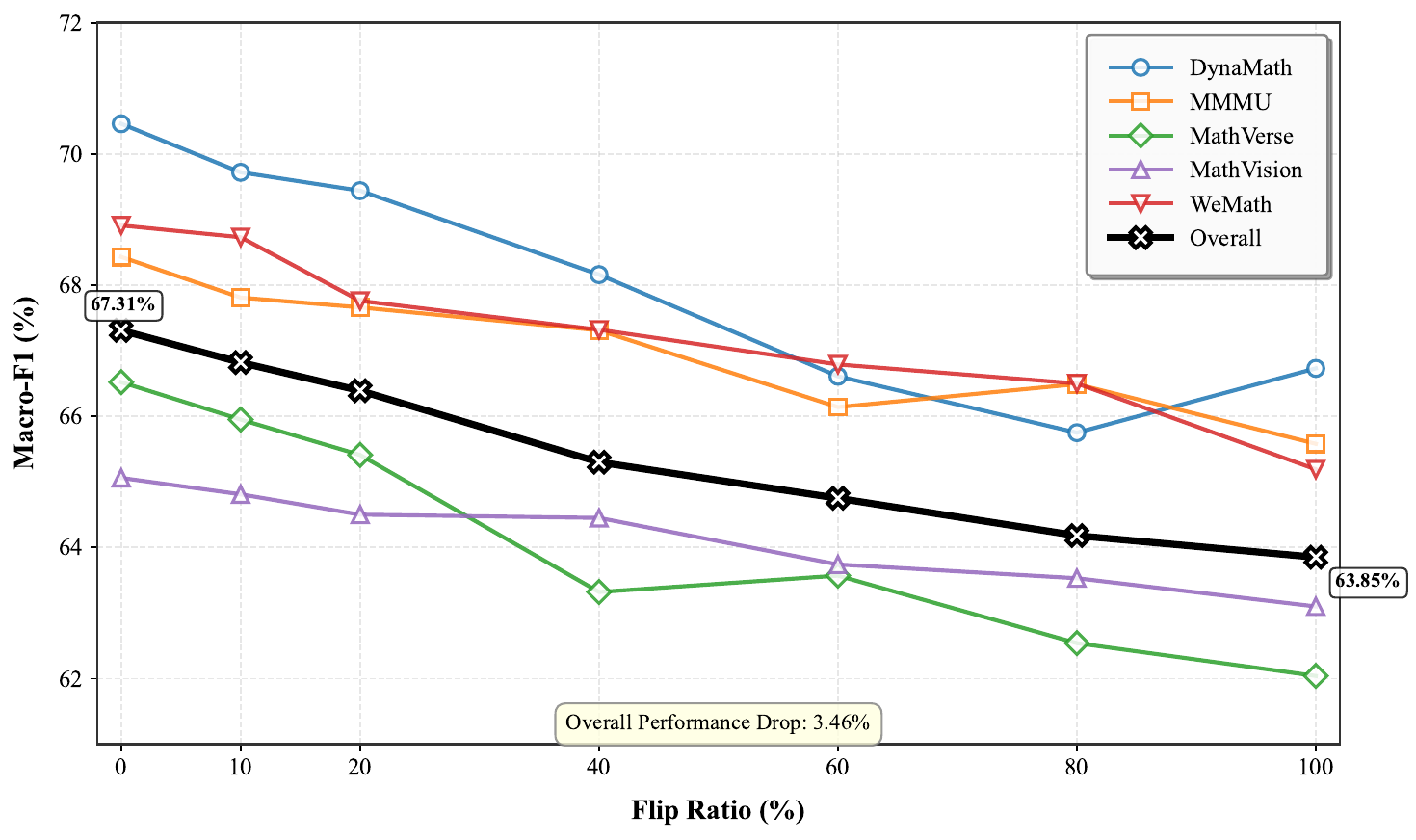}
  \caption{\textbf{Constraint quality--performance causal curves under controlled noise.}}
  \label{fig:causal_curve}
\end{figure}

\definecolor{headerblue}{HTML}{DCEAF7}     
\definecolor{groupblue}{HTML}{EEF3F8}      
\definecolor{fullgreen}{HTML}{E8F4EA}      
\definecolor{overallgray}{HTML}{F5F5F5}    
\definecolor{posgain}{HTML}{1E88E5}        
\definecolor{neggain}{HTML}{C62828}        
\definecolor{bestgreen}{HTML}{2E7D32}      

\sisetup{
  detect-weight=true,
  detect-family=true,
  table-number-alignment=center
}

\begin{table*}[t]
\centering
\caption{\textbf{Key ablations on VisualProcessBench} (Macro-F1; higher is better). $\Delta$ is relative to the full method.}
\label{tab:ablation_vpbench_key}
\small
\setlength{\tabcolsep}{5pt}
\renewcommand{\arraystretch}{1.12}
\resizebox{\textwidth}{!}{%
\begin{tabular}{@{}l
S[table-format=2.2]
S[table-format=2.2]
S[table-format=2.2]
S[table-format=2.2]
S[table-format=2.2]
S[table-format=2.2]
S[table-format=+2.2]
@{}}
\toprule
\rowcolor{headerblue}
\textbf{Variant} &
\textbf{DynaMath} & \textbf{MMMU} & \textbf{MathVerse} &
\textbf{MathVision} & \textbf{WeMath} & \textbf{Overall} &
\textbf{$\Delta$} \\
\midrule

\rowcolor{groupblue}
\multicolumn{8}{@{}l}{\textbf{Full Method}} \\
\rowcolor{fullgreen}
\textbf{Full (EVPV + gating)} &
\bfseries 69.57 & \bfseries 68.86 & \bfseries 67.09 &
\bfseries 65.27 & \bfseries 69.11 & \bfseries 67.46 &
\bfseries \abpos{+0.00} \\
\addlinespace[2pt]

\rowcolor{groupblue}
\multicolumn{8}{@{}l}{\textbf{Evidence / structure ablations}} \\
w/o structured facts (caption-only)     & 67.75 & 58.09 & 63.48 & 60.68 & 67.10 & 63.38 & \abneg{-4.08} \\
w/o constraints (facts = $\varnothing$) & 66.66 & 55.80 & 62.61 & 59.13 & 65.81 & 62.11 & \abneg{-5.35} \\
w/ shuffled facts (structure corrupted) & 62.86 & 52.57 & 59.81 & 58.52 & 64.77 & 59.82 & \abneg{-7.64} \\
\addlinespace[2pt]

\rowcolor{groupblue}
\multicolumn{8}{@{}l}{\textbf{Remove modalities / severe corruption}} \\
w/o vision (text-only judge, keep JSON) & 58.44 & 49.44 & 53.59 & 54.07 & 61.02 & 54.93 & \abneg{-12.53} \\
w/o vision \& w/o JSON (text-only)      & 54.49 & 43.93 & 42.78 & 50.84 & 53.78 & 48.23 & \abneg{-19.23} \\
w/ drop-facts corruption                & 34.90 & 34.40 & 36.29 & 36.14 & 35.96 & 35.77 & \abneg{-31.69} \\
\bottomrule
\end{tabular}%
}
\end{table*}

\subsection{Exp-3: Perception Evidence Quality and Its Causal Impact on Verification}
\label{sec:exp:perception_causal}

EVPV is motivated by a single principle: \emph{reliable visual evidence is a prerequisite for meaningful process verification}. We therefore examine this principle from two complementary angles---\textbf{(i) intervention on the policy’s perceived evidence} and \textbf{(ii) controlled degradation of the verifier’s extracted constraints}---to quantify both the sensitivity of multimodal reasoning to perception and the causal role of constraint fidelity in step verification.

\paragraph{(A) Perception interventions for the policy.}
To measure how strongly multimodal reasoning depends on perception quality, we evaluate the \emph{same} questions under four controlled settings:
(I) \textbf{Normal} (image+$q$),
(II) \textbf{Oracle perception} (image+$q$ plus an oracle structured description),
(III) \textbf{Noisy perception} (image+$q$ plus a corrupted description),
and (IV) \textbf{Text-only} (remove the image).
We run a fixed policy model for all settings and report answer accuracy and PRM trajectory scores (Eq.~\ref{eq:traj_score}).
Table~\ref{tab:attrib} shows two consistent patterns: providing \textbf{oracle perception} substantially improves accuracy, while \textbf{text-only} performance drops sharply, indicating that perception is a dominant bottleneck; moreover, our PRM yields a \textbf{monotonic ordering of trajectory scores} aligned with perception quality: (II)$>$(III)$>$(I)$>$(IV), matching EVPV’s intent that weakened visual evidence should not produce a strong ``correct process'' signal.

\paragraph{(B) Causal curve via constraint corruption.}
EVPV further attributes its gains to the fidelity of the extracted structured constraints used to validate checklist claims.
To test this causally, we inject controlled noise into the constraint set by randomly flipping a fraction of constraint fields (flip ratio), while keeping the policy, verifier/judge, and scoring procedure fixed.
As shown in Figure~\ref{fig:causal_curve}, VisualProcessBench Macro-F1 decreases monotonically as the flip ratio increases across all evaluated judges, providing interventional evidence that verification quality is driven by \textbf{constraint fidelity and premise verification} rather than incidental prompt effects. The mild drop under low noise also indicates that reliability gating is not overly brittle: small constraint errors do not immediately collapse step judgments.

\subsection{Exp-4: Ablation Studies}
\label{sec:exp:ablation}

We ablate core components of EVPV to identify which parts are responsible for the verification and reranking gains. Table~\ref{tab:ablation_vpbench_key} reports representative variants on VisualProcessBench (Macro-F1).

The trends closely match the EVPV design. First, premise verification requires \emph{usable structured evidence}. Replacing structured constraints with caption-only descriptions reduces overall Macro-F1 by 4.08 points, and completely removing constraints (facts = $\varnothing$) further degrades performance (-5.35). This shows that simply having additional text context is insufficient; the verifier benefits from structured, matchable facts that can support checklist claims.When we keep the same facts but shuffle them to corrupt the relational structure, Macro-F1 drops more sharply (-7.64). This indicates that EVPV is not merely exploiting the presence of extra tokens, but relies on faithful entity/relation alignment between checklist items and evidence to compute reliability and gate rewards appropriately.

EVPV still benefits from direct visual input. Making the judge text-only while
keeping JSON constraints reduces Macro-F1 by 12.53, and removing both vision
and JSON drops it by 19.23. Severe evidence loss (drop-facts) collapses
performance by 31.69, indicating that calibration fails when constraints are
too incomplete.

\section{Discussion}
\label{sec:discussion}
EVPV helps by separating two failure sources that standard VL-PRMs often mix:
\emph{bad premises} (misread or hallucinated visual facts) versus \emph{bad
reasoning}. We prompt the policy to state a step-wise visual checklist, verify
those claims against independently extracted \emph{structured} constraints, and
use the resulting reliability to gate rewards. This premise-first design fits
the broader lesson that grounded reasoning depends on faithful perception
\citep{zhang2025mm} and echoes ``generate, then verify'' style faithfulness
checks \citep{wu2025generate}. It is also complementary to stronger/generative
verifiers \citep{she2025r,zhao2025genprm,khalifa2025process}: deeper deliberation
does not fix reasoning built on a wrong visual premise, whereas EVPV explicitly
down-weights rewards when the premise is unreliable.

Viewed another way, EVPV is a visual-specific calibration layer: it tempers
overconfident step rewards under uncertain perception \citep{ye2025uncertainty,park2025know}.
Empirically, it improves step verification on VisualProcessBench
(Table~\ref{tab:vpbench}) and yields stronger Best-of-$N$ reranking across
InternVL2.5 policy sizes (Table~\ref{tab:bon}), with larger gains on
perception-heavy benchmarks. Compared with tool-based approaches such as TIM-PRM
\citep{kuang2025tim}, EVPV trades some evidence granularity for efficiency by
extracting evidence once and reusing it across steps and candidates.

Finally, our corruption and ablation results show that evidence quality matters:
performance drops smoothly as constraints are corrupted (Figure~\ref{fig:causal_curve})
and falls when structured evidence is removed (Table~\ref{tab:ablation_vpbench_key}).
In the larger landscape of grounded verification and process alignment (e.g.,
MJ1/PaLMR-style verified multimodal reasoning), EVPV should be read as a
PRM-focused component: a lightweight test-time calibration layer, not a full
grounded reasoning pipeline.

\section{Conclusion}
\label{sec:conclusion}
We introduced \textbf{Explicit Visual Premise Verification (EVPV)}, a test-time
framework that calibrates multimodal step rewards using premise reliability.
EVPV makes visual premises explicit via a checklist, verifies them against
structured constraints extracted once per instance, and gates rewards for
visually dependent steps. To make this practical in deployment, we trained
\textbf{EVPV-PRM}, a Qwen2.5-VL-Instruct-7B based probabilistic step verifier
whose rewards can be calibrated by EVPV. Across VisualProcessBench and six
downstream benchmarks, EVPV improves step verification and yields overall
Best-of-$N$ reranking gains; controlled constraint corruption leads to monotonic
degradation, supporting the role of evidence fidelity.

\paragraph{Limitations and future work.}
EVPV depends on constraint coverage/accuracy and on checklist completeness.
We currently use a candidate-level (global) reliability signal, which may
propagate a local visual misread to the whole trajectory; step-local reliability
is an important next step. The visual checklist is self-reported by the policy,
so it can under-report visual dependency (e.g., outputting \texttt{null} when a
step actually relies on the image); an external dependency detector/auditor
would improve robustness. Finally, the constraint extractor is trained on
teacher-generated pseudo-gold constraints, so fidelity should be validated with
human-annotated subsets (precision/recall), not only internal matching signals.

\bibliography{conference}
\bibliographystyle{conference}

\appendix



\definecolor{EVPVBlue}{RGB}{55,130,180}
\definecolor{EVPVBlueLight}{RGB}{232,244,252}
\definecolor{EVPVGray}{RGB}{245,245,245}

\newtcolorbox{evpvbox}[1]{%
  enhanced,
  colback=white,
  colframe=SuccessGreen,
  boxrule=1pt,
  arc=3pt,
  outer arc=3pt,
  left=8pt,right=8pt,top=8pt,bottom=8pt,
  title={#1},
  coltitle=white,
  colbacktitle=SuccessGreen,
  fonttitle=\bfseries\small\sffamily,
  attach boxed title to top left={xshift=6mm,yshift=-3mm},
  boxed title style={
    sharp corners,
    boxrule=0pt,
    arc=2pt,
    left=6pt,right=6pt
  },
  shadow={1.5pt}{-1.5pt}{0pt}{black!20},
}

\lstdefinestyle{evpvjson}{%
  basicstyle=\ttfamily\footnotesize,
  columns=fullflexible,
  breaklines=true,
  breakatwhitespace=false,
  showstringspaces=false,
  frame=none,
  xleftmargin=0pt,
  aboveskip=2pt,
  belowskip=0pt
}


\section{Structured Visual Constraint Schema}
\label{app:schema}

The constraint extractor $E_\phi$ maps an image--question pair $(I,q)$ to a
structured set $\mathcal{C}=\{c_k\}_{k=1}^{K}$. Each $c_k$ belongs to one of
three categories: \emph{numeric}, \emph{relation}, or \emph{structure}. The
schema is serialized as a JSON array and is used as the direct supervision
target during SFT (Appendix~\ref{app:training}).

\subsection{Complete Example}
\label{app:schema:example}

The following JSON shows a representative constraint set $\mathcal{C}$ for a
geometry problem whose image depicts a combined cone-and-cylinder solid with
labeled dimensions.

At test time, $E_\phi$ predicts $\mathcal{C}$ from $(I,q)$ directly; no gold
constraints are used. During training (Appendix~\ref{app:training}), the
teacher model provides $\mathcal{C}^\star$ as supervision targets.

\begin{figure*}[!b]
  \centering
  \begin{minipage}{0.95\textwidth}
    \begin{evpvbox}{Example: structured visual constraint set $\mathcal{C}$}
\lstset{style=evpvjson}
\begin{lstlisting}
[
  {
    "category": "numeric",
    "entity": "cylinder base radius",
    "attribute": "length",
    "value": 3,
    "unit": "cm",
    "confidence": 0.95
  },
  {
    "category": "numeric",
    "entity": "cylinder height",
    "attribute": "length",
    "value": 8,
    "unit": "cm",
    "confidence": 0.92
  },
  {
    "category": "numeric",
    "entity": "cone height",
    "attribute": "length",
    "value": 4,
    "unit": "cm",
    "confidence": 0.88
  },
  {
    "category": "relation",
    "type": "equal",
    "entities": ["cone base radius", "cylinder base radius"],
    "direction": null,
    "confidence": 0.97
  },
  {
    "category": "structure",
    "type": "composite",
    "parts": ["cylinder", "cone"],
    "attachment": ["cone placed on top of cylinder"],
    "adjacency": [],
    "confidence": 0.94
  }
]
\end{lstlisting}
    \end{evpvbox}
  \end{minipage}
  \caption{A representative structured constraint set $\mathcal{C}$ serialized as JSON.}
  \label{fig:app_schema_json}
\end{figure*}

\subsection{Schema Specification}
\label{app:schema:spec}
\definecolor{headerblue}{HTML}{DCEAF7}     
\definecolor{numericrow}{HTML}{EEF6FF}     
\definecolor{relationrow}{HTML}{F3F0FA}    
\definecolor{structurerow}{HTML}{EEF8F0}   
\definecolor{codegray}{HTML}{F7F7F7}       

\begin{table*}[t]
\centering
\caption{\textbf{Top-level fields for each constraint category.}
\textsuperscript{*}\texttt{confidence} is a model-estimated reliability weight in $[0,1]$ and is used during matching (Appendix~\ref{app:matching}).}
\label{tab:schema}
\small
\renewcommand{\arraystretch}{1.22}
\setlength{\tabcolsep}{7pt}
\begin{tabularx}{\textwidth}{@{}p{2.4cm} p{4.6cm} X@{}}
\toprule
\rowcolor{headerblue}
\textbf{Category} & \textbf{Key fields} & \textbf{Description} \\
\midrule

\rowcolor{numericrow}
\texttt{numeric}
  & \texttt{entity}, \texttt{attribute}, \texttt{value}, \texttt{unit}, \texttt{confidence}\textsuperscript{*}
  & A measurable fact associated with a named visual entity.
    \texttt{entity} is a label or description of the object (e.g., \texttt{"segment AB"});
    \texttt{attribute} names the quantity (e.g., \texttt{"length"}, \texttt{"angle"}, \texttt{"count"});
    \texttt{value} is a numeric literal; \texttt{unit} is optional (e.g., \texttt{"cm"}, \texttt{"degrees"}). \\
\midrule

\rowcolor{relationrow}
\texttt{relation}
  & \texttt{type}, \texttt{entities}, \texttt{direction}, \texttt{confidence}
  & A geometric or logical relationship between two or more entities.
    \texttt{type} encodes one of
    \{\texttt{parallel}, \texttt{perpendicular}, \texttt{equal},
      \texttt{subset}, \texttt{incident}, \texttt{adjacent},
      \texttt{greater}, \texttt{less}\}.
    \texttt{entities} is an ordered list of participants; \texttt{direction} is optional (e.g., \texttt{"AB$\to$CD"}). \\
\midrule

\rowcolor{structurerow}
\texttt{structure}
  & \texttt{type}, \texttt{parts}, \texttt{attachment}, \texttt{adjacency}, \texttt{confidence}
  & Compositional or topological description of a multi-part figure.
    \texttt{type} is one of \{\texttt{composite}, \texttt{graph}, \texttt{table}, \texttt{sequence}\}.
    \texttt{parts} lists sub-components; \texttt{attachment} and \texttt{adjacency} are optional relational lists specifying how parts connect. \\
\bottomrule
\end{tabularx}
\end{table*}

\section{Checklist--Constraint Matching Function}
\label{app:matching}

We describe the type-aware matching function $m(v_j,\mathcal{C})$ that maps a
single checklist claim $v_j$ to a support score $p_j\in[0,1]$. This matching
is used to compute the candidate-level visual reliability $r$ in
Eq.~\ref{eq:geomagg} of the main paper.

\subsection{Claim Parsing}
Each checklist item $v_j$ (produced by the policy's \texttt{visualdependency}
field) is a short natural-language assertion. We classify it into one of three
claim types---\emph{numeric}, \emph{relational}, or \emph{structural}---using a
lightweight classifier trained on the schema vocabulary.

\paragraph{Unclassifiable claims.}
If a claim cannot be reliably parsed into the schema (e.g., too vague or
out-of-domain), we assign a \emph{neutral} support score $p_j=0.5$. This
represents uncertainty rather than contradiction and prevents the reliability
score from collapsing due to parser limitations.

\subsection{Type-Specific Matching}

\paragraph{Preliminaries (tokenization and similarity).}
We normalize strings by lowercasing, removing punctuation, and splitting on
whitespace. For any string $x$, let $\mathrm{Tok}(x)$ be its token set.
We use Jaccard similarity between token sets:
\[
\mathrm{Jaccard}(A,B)=\frac{|A\cap B|}{|A\cup B|}.
\]
We treat two entity strings as \emph{approximately matched} if
$\mathrm{Jaccard}(\mathrm{Tok}(e),\mathrm{Tok}(e')) \ge 0.5$.

\paragraph{Numeric matching.}
For a numeric claim asserting ``entity $e$ has attribute $a$ equal to value $x$
(unit $u$)'', we search $\mathcal{C}$ for numeric constraints $c_k$ with
\texttt{attribute}$=a$ and \texttt{entity}$\approx e$. Among all matched
constraints, we choose the one with highest \texttt{confidence} and compute
\begin{equation}
p_j^{\text{num}}
  = \mathbb{I}\!\left[\frac{\lvert x-c_k.\text{value}\rvert}{\max(\lvert x\rvert,1)}<\delta\right]\cdot c_k.\text{confidence},
\label{eq:numeric_match}
\end{equation}
with tolerance $\delta=0.15$.

\paragraph{No match for a well-formed claim.}
If the claim is parsed successfully as numeric but \emph{no} numeric constraint
matches its entity/attribute, we set $p_j^{\text{num}}=0$. This corresponds to
``unsupported by extracted evidence'' and is intended to penalize hallucinated
or misread premises under the geometric-mean aggregation (Eq.~\ref{eq:geomagg}).

\paragraph{Relation matching.}
For a relational claim asserting a relation type $t$ over entities
$\{e_1,\ldots,e_n\}$, we search $\mathcal{C}$ for relation constraints with
\texttt{type}$=t$. For each candidate constraint $c_k$, we compute entity-set
overlap by comparing the \emph{union of tokens}:
\[
E_{\text{claim}}=\bigcup_{i=1}^{n}\mathrm{Tok}(e_i),\qquad
E_{\text{con}}=\bigcup_{e'\in c_k.\text{entities}}\mathrm{Tok}(e').
\]
We define
\begin{equation}
p_j^{\text{rel}}
  = \max_{c_k\in\mathcal{C}^{(t)}}
    \mathrm{Jaccard}(E_{\text{claim}},E_{\text{con}})\cdot c_k.\text{confidence},
\label{eq:rel_match}
\end{equation}
where $\mathcal{C}^{(t)}$ is the subset of constraints with \texttt{type}$=t$.
Synonym groups are used to handle equivalent relation labels (e.g.,
\texttt{perpendicular} $\leftrightarrow$ \texttt{orthogonal}).

\paragraph{Relation no-match.}
If the claim is parsed as a relation but $\mathcal{C}^{(t)}$ is empty or the
maximum overlap is $0$, we set $p_j^{\text{rel}}=0$.

\paragraph{Structural matching.}
For a structural claim specifying a set of parts $P=\{p_1,\ldots,p_m\}$, we
search $\mathcal{C}$ for structure constraints of compatible \texttt{type}
(e.g., \texttt{composite}, \texttt{graph}, \texttt{table}, \texttt{sequence}).
Let $\mathrm{TokSet}(P)=\bigcup_{p\in P}\mathrm{Tok}(p)$ and similarly for
$c_k.\text{parts}$. We compute:
\begin{equation}
p_j^{\text{str}}
  = \max_{c_k\in\mathcal{C}^{\text{struct}}}
    \mathrm{Jaccard}\!\big(\mathrm{TokSet}(P), \mathrm{TokSet}(c_k.\text{parts})\big)\cdot c_k.\text{confidence}.
\label{eq:struct_match}
\end{equation}
If no structure constraint exists or overlap is $0$, we set $p_j^{\text{str}}=0$.

\subsection{Final Per-Claim Score $p_j$}
The per-claim score $p_j$ is the type-specific score from the matched routine:
\[
p_j=
\begin{cases}
p_j^{\text{num}}, & \text{if } v_j \text{ is numeric},\\
p_j^{\text{rel}}, & \text{if } v_j \text{ is relational},\\
p_j^{\text{str}}, & \text{if } v_j \text{ is structural},\\
0.5, & \text{if } v_j \text{ is unclassifiable}.
\end{cases}
\]
Note that ``no match'' for a \emph{well-formed} claim yields $p_j=0$, while
``cannot parse'' yields a neutral $p_j=0.5$.

\subsection{Reliability Score $r$}
Given $M$ checklist claims with support scores $\{p_j\}_{j=1}^{M}$, we compute
the candidate-level visual reliability score $r$ using the smoothed geometric
mean (same as Eq.~\ref{eq:geomagg} in the main paper):
\begin{equation}
r=\exp\!\left(\frac{1}{M}\sum_{j=1}^{M}\log(\epsilon+p_j)\right),
\quad \epsilon=10^{-6}.
\label{eq:geo_mean_app}
\end{equation}
This aggregation is deliberately sensitive to catastrophic premise failures:
if any well-formed claim is clearly unsupported ($p_j\approx 0$), then $r$
drops sharply and reliability gating attenuates the rewards for visually
dependent steps.

\section{Training Details}
\label{app:training}

\subsection{Dataset Construction and Leakage Control}

\paragraph{Train/eval boundary.}
We use \textbf{VisualProcessBench} \citep{wang2025visualprm} \emph{only} for
evaluation. No VisualProcessBench instances (images or questions) are used in
training either $E_\phi$ or $V_\theta$. All training data are sampled from
\textbf{VisualPRM400K} \citep{wang2025visualprm}.
We further restrict training to geometry- and table-focused sources within
VisualPRM400K: \textsc{Geo170K}, \textsc{GeometryData}, \textsc{GeomVerse},
\textsc{GEOS}, \textsc{MAVIS-Geometry}, \textsc{TabMWP}, and \textsc{UniGeo}.
This design prevents overlap with VisualProcessBench by construction and keeps
the benchmark as a held-out testbed.

\paragraph{Constraint distillation corpus (for $E_\phi$).}
We construct an SFT corpus of 26{,}454 image--question pairs from the above
VisualPRM400K subsets. For each instance, we use
\texttt{qwen3-vl-235b-a22b-instruct} as a teacher to produce a pseudo-gold
constraint set $\mathcal{C}^\star$.
The teacher is prompted with (i) the schema in Appendix~\ref{app:schema} and
(ii) the reference solution steps for the problem, and is instructed to output
a JSON array of constraints that are directly supported by the image and
solution-critical. Responses that fail schema validation are filtered. These
pairs form the SFT dataset for $E_\phi$.

\paragraph{Preference data for DPO (for $E_\phi$).}
To improve fidelity on hard cases, we additionally sample 5{,}832 instances and
ask the same teacher to generate rewritten constraint variants that form a
preferred/rejected pair $(C^+,C^-)$. These preference pairs are used in a DPO
stage after SFT (details below).

\paragraph{Step verifier training corpus (for $V_\theta$).}
We train $V_\theta$ on 19{,}490 step-labeled trajectories sampled from
VisualPRM400K. Each trajectory provides step-level supervision derived from its
reference trace, producing binary labels $y_t\in\{0,1\}$. We emphasize that
these labels are \emph{not} taken from VisualProcessBench; VisualProcessBench
labels are used only for evaluation.

\subsection{Constraint Extractor $E_\phi$}

\paragraph{Architecture.}
$E_\phi$ is initialized from a pre-trained multimodal VLM backbone
(Qwen2.5-VL-Instruct-7B) and fine-tuned to generate structured constraint JSON
conditioned on $(I,q)$.

\paragraph{SFT stage.}
We minimize the next-token prediction loss on the JSON serialization of
$\mathcal{C}^\star$:
\begin{equation}
\mathcal{L}_{\text{con}}(\phi)=-\log P_\phi(\mathcal{C}^\star\mid I,q).
\end{equation}
We train on 26{,}454 instances using AdamW with learning rate $2\times 10^{-5}$,
linear warmup over the first 3\% of steps, cosine decay, batch size 16, and 3
epochs. Maximum sequence length is 4096 tokens.

\paragraph{DPO stage.}
After SFT, we apply DPO using 5{,}832 preference pairs. For each instance, we
optimize:
\begin{equation}
\mathcal{L}_{\text{DPO}}(\phi)
  = -\log \sigma\!\left(
      \beta_{\text{dpo}}
      \left[\log P_\phi(C^+\mid I,q)-\log P_\phi(C^-\mid I,q)\right]
    \right),
\end{equation}
with $\beta_{\text{dpo}}=0.1$ and preference-pair weight
$\lambda_{\text{dpo}}=0.1$. DPO runs for 1 epoch with learning rate
$5\times 10^{-6}$.
Preferred/rejected pairs are produced by the teacher via schema-preserving
rewrites; the selection criterion is a schema-aware distance to the teacher's
pseudo-gold constraints (Appendix~\ref{app:matching}).

\subsection{Step Verifier $V_\theta$}

$V_\theta$ is fine-tuned from the same Qwen2.5-VL-Instruct-7B backbone using binary
cross-entropy on step-level correctness labels from the VisualPRM400K-derived
training corpus (19{,}490 trajectories):
\begin{equation}
\mathcal{L}_V(\theta)
  = -\sum_{t=1}^{T}\left[y_t\log u_t+(1-y_t)\log(1-u_t)\right],
\end{equation}
where $u_t=P_\theta(y_t=1\mid I,q,s_{\le t})$.
Training uses AdamW with learning rate $2\times 10^{-5}$, batch size 8, 3 epochs,
and maximum sequence length 8{,}192 tokens.
\emph{Reliability gating is applied only at inference time} as a calibration
layer; the verifier is trained on raw step labels without gating.

\subsection{Reliability Gating Hyperparameters}

The gating factor $\alpha(r)=\sigma(\beta(r-\tau))$ (Eq.~(9) of the main paper)
is controlled by two hyperparameters:
\begin{itemize}
  \item $\tau=0.5$: reliability threshold below which rewards are attenuated.
        A claim-set where every claim is half-supported yields $r\approx 0.5$,
        which maps to $\alpha\approx 0.5$ under our sigmoid.
  \item $\beta=10$: sigmoid sharpness. At $\beta=10$ the transition from near-zero
        attenuation ($r>0.7$) to near-full attenuation ($r<0.3$) spans roughly
        0.4 units of $r$, providing a smooth but decisive gate.
\end{itemize}

\paragraph{Sensitivity analysis.}
Table~\ref{tab:tau_sensitivity} reports VisualProcessBench overall Macro-F1 under
five choices of $\tau$ (with $\beta=10$ fixed). Performance is relatively stable
for $\tau\in[0.4,0.6]$.

\begin{table}[t]
\centering
\caption{VisualProcessBench overall Macro-F1 (\%) under varying reliability threshold $\tau$ ($\beta=10$ fixed).}
\label{tab:tau_sensitivity}
\setlength{\tabcolsep}{6pt}
\begin{tabular}{@{}lccccc@{}}
\toprule
$\tau$ & 0.3 & 0.4 & \textbf{0.5} & 0.6 & 0.7 \\
\midrule
Macro-F1 & 66.91 & 67.23 & \textbf{67.46} & 67.18 & 66.74 \\
\bottomrule
\end{tabular}
\end{table}


\section{Complete Prompt Templates}
\label{app:prompts}

We provide the verbatim prompts used in our pipeline. Placeholders are shown
in angle brackets (\texttt{\{...\}}). Importantly, EVPV relies on
\textbf{structured constraints} (Appendix~\ref{app:schema}) and a
\textbf{type-aware matching function} (Appendix~\ref{app:matching}) to compute
visual reliability; it does \emph{not} require natural-language ``gold image
descriptions'' or an LLM-based checklist auditor at test time.

For clarity, we separate prompts by module:
(i) the \textbf{policy} prompt (producing steps and the visual checklist),
(ii) the \textbf{teacher} prompt used \emph{only for distilling} structured
constraints for training the constraint extractor, and
(iii) an \textbf{optional external step-judge} prompt used when EVPV is plugged
into black-box VLM/LLM judges. Our main EVPV-PRM results use the trained step
verifier $V_\theta$ (probabilistic output) and therefore do not require (iii).

\definecolor{EVPVBlue}{RGB}{55,130,180}

\lstdefinestyle{evpvprompt}{%
  basicstyle=\ttfamily\footnotesize,
  columns=fullflexible,
  breaklines=true,
  breakatwhitespace=false,
  showstringspaces=false,
  frame=none,
  xleftmargin=0pt,
  aboveskip=2pt,
  belowskip=0pt
}

\newtcolorbox{evpvpromptbox}[1]{%
  enhanced,
  colback=LightGray,
  colframe=PrimaryBlue,
  boxrule=1pt,
  arc=3pt,
  outer arc=3pt,
  left=10pt,right=10pt,top=10pt,bottom=10pt,
  title={#1},
  coltitle=white,
  colbacktitle=PrimaryBlue,
  fonttitle=\bfseries\small\sffamily,
  attach boxed title to top left={xshift=8mm,yshift=-3mm},
  boxed title style={
    sharp corners,
    boxrule=0pt,
    arc=2pt,
    left=8pt,right=8pt
  },
  shadow={2pt}{-2pt}{0pt}{black!30},
}

\subsection{Policy Inference Prompt (Steps + Visual Checklist)}
\label{app:prompt_policy}

Used to elicit structured, step-by-step solutions with per-step
\texttt{visualdependency} annotations (our visual checklist) from the policy.
A unique \texttt{nonce} and \texttt{variant\_id} are injected per candidate to
promote diversity across the $N=8$ samples.

\begin{figure*}[t]
  \centering
  \begin{minipage}{1\textwidth}
  \begin{evpvpromptbox}{Policy inference prompt (USER turn + image)}
  \lstset{style=evpvprompt}
\begin{lstlisting}
You are a meticulous and precise AI assistant, an expert in visual
mathematical reasoning. Your primary goal is to solve the user's query by
providing a detailed, step-by-step thought process.

You MUST provide your entire response in a single, valid JSON code block.
Do not include any text, explanations, or markdown formatting outside of the JSON object.

---

### DIVERSITY REQUIREMENTS (VERY IMPORTANT)
- This is reasoning variant #{variant_id}. Your reasoning path should be
  meaningfully different from other variants.
- Try a different logical decomposition, use different intermediate variables,
  or vary the order of non-dependent steps.
- Use this nonce strictly as a randomness anchor for this specific generation:
  {nonce}

---

### JSON OUTPUT SPECIFICATION (CRITICAL)

Your entire output must conform to this JSON schema:

{
  "reasoningprocess": [
    {
      "steptext": "A single, clear step of reasoning...",
      "visualdependency": "A specific, observable fact from the image, or null."
    }
  ],
  "finalanswer": "The final answer."
}

Field-Specific Rules:
1) reasoningprocess (List of Objects):
   - steptext: Each step should represent a single calculation, observation, or deduction.
   - visualdependency (String or null): Include a description if the step directly reads
     a value/label/relation/structure from the image. Use null ONLY for purely abstract steps.
     CRITICAL: use the JSON literal null, NEVER an empty string "".
   - Make the visualdependency statement minimally checkable (one claim per step).
     Examples:
       - "The table entry in row 2, column B is 12."
       - "AB is perpendicular to CD."
       - "The figure is a cylinder with a cone attached on top."
2) finalanswer (String): For multiple-choice output the option letter only (e.g., "A");
   for open-ended output the numerical result only.

---

### USER QUERY
{user_query}
\end{lstlisting}
  \end{evpvpromptbox}
  \end{minipage}
  \caption{Policy prompt used to generate step-by-step reasoning traces with explicit \texttt{visualdependency} checklist items.}
  \label{fig:app_prompt_policy}
\end{figure*}

\subsection{Teacher Prompt for Constraint Distillation (Training Only)}
\label{app:prompt_teacher_constraints}

This prompt is used \emph{only during data construction} to distill pseudo-gold
structured constraints $C^\star$ for training the constraint extractor $E_\phi$.
At test time, EVPV uses the constraints predicted by $E_\phi$ and does not use
any natural-language ``gold description''.

\begin{figure*}[t]
  \centering
  \begin{minipage}{1\textwidth}
  \begin{evpvpromptbox}{Constraint distillation prompt (USER turn + image)}
  \lstset{style=evpvprompt}
\begin{lstlisting}
You are a top-tier image analyst for multimodal math problems.
Given the image and question, extract solution-critical visual facts and output
them as a STRICT JSON array of constraints.

You MUST follow this schema:

Each array element is an object with:
- "category": one of ["numeric", "relation", "structure"]
- "confidence": a float in [0,1]

If category == "numeric", include:
- "entity": string (e.g., "segment AB", "table row 2 col B", "cylinder height")
- "attribute": string (e.g., "length", "angle", "count", "value")
- "value": number
- "unit": string or null (e.g., "cm", "deg", null)

If category == "relation", include:
- "type": one of ["parallel","perpendicular","equal","incident","subset","adjacent","greater","less"]
- "entities": array of strings (2 or more)
- "direction": string or null

If category == "structure", include:
- "type": one of ["composite","graph","table","sequence"]
- "parts": array of strings
- "attachment": array of strings
- "adjacency": array of strings

Rules:
- Only include facts that are directly supported by the image (do not infer hidden values).
- Prefer atomic, checkable facts. Avoid long paragraphs.
- Output ONLY the JSON array. No extra text.

---

Question text:
{question_text}

Image: [image token]
\end{lstlisting}
  \end{evpvpromptbox}
  \end{minipage}
  \caption{Teacher prompt used to distill pseudo-gold structured constraints $C^\star$ for training $E_\phi$.}
  \label{fig:app_prompt_teacher_constraints}
\end{figure*}

\subsection{Constraint Extractor Inference Prompt (If Using Prompted Decoding)}
\label{app:prompt_extractor_infer}

In our main system, $E_\phi$ is a fine-tuned model trained to directly generate
constraints in the JSON format (Appendix~\ref{app:schema}). If one instantiates
$E_\phi$ via prompted decoding (e.g., for ablations), we use the following prompt.

\begin{figure*}[t]
  \centering
  \begin{minipage}{1\textwidth}
  \begin{evpvpromptbox}{Constraint extractor inference prompt (USER turn + image)}
  \lstset{style=evpvprompt}
\begin{lstlisting}
Extract solution-critical visual facts from the image and output a STRICT JSON
array of constraints following the schema below.

Schema (same as training):
- category in ["numeric","relation","structure"]
- confidence in [0,1]
- numeric: {entity, attribute, value, unit, confidence}
- relation: {type, entities, direction, confidence}
- structure: {type, parts, attachment, adjacency, confidence}

Output ONLY the JSON array. No other text.

Question:
{question_text}

Image: [image token]
\end{lstlisting}
  \end{evpvpromptbox}
  \end{minipage}
  \caption{Optional inference-time prompt for generating constraints in JSON form (used only when $E_\phi$ is instantiated via prompting).}
  \label{fig:app_prompt_extractor_infer}
\end{figure*}

\subsection{Optional External Step-Judge Prompt (EVPV Plug-in)}
\label{app:prompt_external_judge}

EVPV is judge-agnostic: it can calibrate rewards from an external black-box
judge that outputs binary step decisions. This prompt is used \emph{only} for
plug-in experiments where the base reward is provided by an external VLM/LLM
(e.g., GPT/Gemini). Our main EVPV-PRM results use a trained step verifier
$V_\theta$ (probabilistic output) and thus do not require this prompt.

\begin{figure*}[t]
  \centering
  \begin{minipage}{1\textwidth}
  \begin{evpvpromptbox}{External step-judge prompt (USER turn + image)}
  \lstset{style=evpvprompt}
\begin{lstlisting}
You are a professional expert in mathematical reasoning.
You will judge whether the CURRENT solution step is correct given the image,
the problem, and the previous steps.

Output format: a STRICT JSON object with exactly one key "judgment",
whose value is an integer 1 or -1.
- 1 means the step is correct.
- -1 means the step is incorrect (contradicts the image/question/previous steps,
  or is invalid reasoning).
Do NOT output any other text or explanation.

Problem:
{question_text}

Structured constraints (evidence C):
{constraints_json_text}

Previous steps:
{history_steps_text}

CURRENT step to evaluate:
{current_step_text}

Problem image: [image token]
\end{lstlisting}
  \end{evpvpromptbox}
  \end{minipage}
  \caption{Optional prompt for external step judges in the EVPV plug-in setting. The judge's output is treated as $R_t^{\mathrm{base}}\in\{-1,+1\}$ and then calibrated by EVPV gating.}
  \label{fig:app_prompt_external_judge}
\end{figure*}

\subsection{Step Error Attribution in VisualProcessBench}
\label{app:vpbench_error_attrib}

\paragraph{Step-level evaluation labels (VisualProcessBench).}
We use the process-level correctness annotations from VisualProcessBench
\citep{wang2025visualprm}, which provides per-step labels $\{y_t\}$
($y_t\in\{0,1\}$) for each solution trace. \textbf{These labels are used only for
evaluation} of step verification (e.g., Macro-F1 in Table~\ref{tab:vpbench}) and
for the error analysis below. The step verifier $V_\theta$ is trained on
VisualPRM400K-derived step-labeled trajectories as described in
Appendix~\ref{app:training}, not on VisualProcessBench.

\paragraph{Step-level error-type attribution in VisualProcessBench.}
VisualProcessBench provides step-level correctness labels ($+1$ = correct,
$-1$ = incorrect) for each solution trace. To understand \emph{why} incorrect
steps fail and to support the error-distribution statistics reported in the
main paper (e.g., the pie charts), we performed \textbf{error-type classification}
on all steps marked incorrect ($-1$).

The taxonomy is two-level. \textbf{Top-level categories}:
\textit{Visual Misinterpretation} (misreading or misusing the image),
\textit{Logical Error} (invalid deduction or reasoning chain),
\textit{Calculation Error} (arithmetic or algebraic mistake),
\textit{Knowledge Error} (wrong formula or domain fact), and
\textit{Incompleteness} (step is underspecified or missing key detail).
\textit{Visual Misinterpretation} is further split into \textbf{sub-types}:
\textit{Structural Misunderstanding} (wrong spatial or geometric structure),
\textit{Value Misreading} (wrong number or measure from the figure), and
\textit{Object Misidentification} (wrong object, label, or correspondence).

We used a dedicated prompt (below) with \textbf{Gemini-2.5-Pro} to assign, for
each incorrect step, one top-level category and, when applicable, one visual
sub-type. The model was given the problem text, the image, the full solution,
and the index of the incorrect step. \textbf{Human annotators} then reviewed a
subset of model-predicted labels, correcting misclassifications. Disagreements
were resolved by discussion or a third annotator. Statistics reported in the
main paper are computed from the final, human-verified distribution over all
incorrect steps.

\begin{figure*}[t]
  \centering
  \begin{minipage}{1\textwidth}
  \begin{evpvpromptbox}{Error-type classification prompt (Gemini-2.5-Pro)}
  \lstset{style=evpvprompt}
\begin{lstlisting}
Task. You are an expert in mathematical reasoning and multimodal evaluation.
You will be given a math problem, an image, a step-by-step solution, and the
index of one step that is already known to be incorrect. Your job is to
classify the type of error that best explains why this step is wrong.

Top-level error types (choose exactly one):
- Visual Misinterpretation -- The step is wrong because it misreads or misuses
  information from the image (wrong shape, number, label, relation, or structure).
- Logical Error -- The step is wrong due to invalid deduction, wrong implication,
  or broken reasoning chain (not primarily a visual or calculation mistake).
- Calculation Error -- The step applies correct reasoning but contains an arithmetic
  or algebraic mistake.
- Knowledge Error -- The step uses a wrong formula, definition, or domain fact.
- Incompleteness -- The step is underspecified, skips necessary detail, or does not
  fully justify the conclusion.

If you choose Visual Misinterpretation, also choose exactly one sub-type:
- Structural Misunderstanding -- Wrong spatial, geometric, or compositional structure.
- Value Misreading -- Wrong numeric value or measure read from the figure.
- Object Misidentification -- Wrong object, label, or correspondence.

Output format. Reply with a single JSON object:
{
  "top_level": "Visual Misinterpretation" | "Logical Error" |
               "Calculation Error" | "Knowledge Error" | "Incompleteness",
  "visual_subtype": "Structural Misunderstanding" | "Value Misreading" |
                    "Object Misidentification" | null
}
Set "visual_subtype" to null if "top_level" is not "Visual Misinterpretation".

Input.
Problem: {question_text}
Image: [image]
Solution steps:
Step 1: ...
Step 2: ...
...
The following step is INCORRECT (index {step_index}): ...
Classify the error type for this step.
\end{lstlisting}
  \end{evpvpromptbox}
  \end{minipage}
  \caption{Prompt used for error-type classification of incorrect steps in VisualProcessBench.}
  \label{fig:app_prompt_error_type}
\end{figure*}


\section{Alternative Score Aggregation Strategies}
\label{app:aggregation}

The main paper uses the \textbf{geometric-mean trajectory score} in
Eq.~\ref{eq:traj_score} for Best-of-$N$ reranking. This choice is deliberate:
reliability gating (Eq.~\ref{eq:gated_reward}) rescales step-reward magnitudes,
and a magnitude-sensitive aggregation ensures that gating can affect candidate
ranking. Here we report results under five aggregation strategies implemented
in our evaluation pipeline.

\paragraph{Aggregation strategies.}
Let $\{R_t\}_{t=1}^{T}$ denote the (gated) step rewards for a candidate solution,
where $R_t\in[-1,1]$ for our probabilistic PRM judge and $R_t\in\{-1,+1\}$ for
binary external judges. We consider:

\begin{enumerate}
  \item \textbf{Geometric Mean (main paper).}
  We first map rewards to positive values
  $\tilde{R}_t=\epsilon+\frac{R_t+1}{2}$ and compute
  \[
    \mathrm{Score}(S)
    =\exp\!\left(\frac{1}{T}\sum_{t=1}^{T}\log \tilde{R}_t\right),
  \]
  which is identical to Eq.~\ref{eq:traj_score}. This aggregation is sensitive
  to any low-scoring step, matching the intuition that a single catastrophic
  premise failure can invalidate an entire trace.

  \item \textbf{Correctness Rate (alternative).}
  \[
    \mathrm{Score}(S)=\frac{1}{T}\sum_{t=1}^{T}\mathbb{I}[R_t>0].
  \]
  This sign-based aggregation is simple but largely insensitive to magnitude
  rescaling, and thus can under-utilize reliability gating.

  \item \textbf{Streak Score (alternative).}
  We reward consecutive correct-step runs: the score is incremented by the
  current streak length on each correct step and decremented by 1 on each
  incorrect step, then normalized to $[0,1]$.

  \item \textbf{Weighted Correctness (alternative).}
  Later steps receive linearly higher weight. Let $w_t=t$ and compute
  \[
    \mathrm{Score}(S)=\frac{\sum_{t=1}^{T} w_t R_t - W_{\min}}{W_{\max}-W_{\min}},
  \]
  where $W_{\max/\min}$ are the maximum/minimum achievable weighted sums.

  \item \textbf{First-Error Position (alternative).}
  \[
    \mathrm{Score}(S)=\frac{i^\ast}{T},
  \]
  where $i^\ast$ is the index of the first step with $R_t<0$; it equals $1.0$
  if no error occurs.
\end{enumerate}

Tables~\ref{tab:agg_8b}--\ref{tab:agg_38b} report Pass@1 and BoN@8 accuracy (\%)
for each strategy across three InternVL2.5 policy scales, where
$\Delta_8=\mathrm{BoN@8}-\mathrm{Pass@1}$.

\noindent
Overall, the geometric mean achieves the best or near-best BoN@8 across scales
and benchmarks while remaining simple to compute. Weighted Correctness is the
most conservative, often over-penalizing candidates with a single minor error.
Correctness Rate and First-Error Position generally track the geometric mean,
indicating that reranking gains are robust to the choice of aggregation, though
magnitude-sensitive aggregations tend to better reflect reliability gating.
\definecolor{headerblue}{HTML}{DCEAF7}      
\definecolor{subheaderblue}{HTML}{EEF4FB}   
\definecolor{bestrowgreen}{HTML}{E8F4EA}    
\definecolor{groupgray}{HTML}{F8F8F8}       
\definecolor{posgain}{HTML}{1E88E5}         
\definecolor{textgray}{HTML}{555555}

\newcommand{\aggimp}[1]{\textcolor{posgain}{\textbf{#1}}}

\begin{table*}[t]
\centering
\caption{\textbf{Best-of-8 reranking under five aggregation strategies, InternVL2.5-8B policy.}
Pass@1 is the same across strategies; BoN@8 and $\Delta_8$ vary.}
\label{tab:agg_8b}
\small
\setlength{\tabcolsep}{3.6pt}
\renewcommand{\arraystretch}{1.14}
\resizebox{\textwidth}{!}{%
\begin{tabular}{l ccc ccc ccc ccc ccc ccc}
\toprule
\rowcolor{headerblue}
& \multicolumn{3}{c}{\textbf{MathVista}}
& \multicolumn{3}{c}{\textbf{MathVision}}
& \multicolumn{3}{c}{\textbf{MathVerse-VO}}
& \multicolumn{3}{c}{\textbf{WeMath}}
& \multicolumn{3}{c}{\textbf{LogicVista}}
& \multicolumn{3}{c}{\textbf{Overall}} \\
\rowcolor{subheaderblue}
\textbf{Strategy}
& \textbf{P@1} & \textbf{B@8} & \textbf{$\Delta_8$}
& \textbf{P@1} & \textbf{B@8} & \textbf{$\Delta_8$}
& \textbf{P@1} & \textbf{B@8} & \textbf{$\Delta_8$}
& \textbf{P@1} & \textbf{B@8} & \textbf{$\Delta_8$}
& \textbf{P@1} & \textbf{B@8} & \textbf{$\Delta_8$}
& \textbf{P@1} & \textbf{B@8} & \textbf{$\Delta_8$} \\
\midrule
\rowcolor{bestrowgreen}
Geometric Mean
& 64.5 & \textbf{76.3} & \aggimp{+11.8}
& 17.0 & \textbf{22.1} & \aggimp{+5.1}
& 22.8 & \textbf{29.5} & \aggimp{+6.7}
& 23.5 & \textbf{37.5} & \aggimp{+14.0}
& 36.4 & \textbf{45.3} & \aggimp{+8.9}
& 32.8 & \textbf{41.7} & \aggimp{+8.9} \\
Correctness Rate
& 64.5 & 75.1 & +10.6
& 17.0 & 21.4 & +4.4
& 22.8 & 28.9 & +6.1
& 23.5 & 36.8 & +13.3
& 36.4 & 44.6 & +8.2
& 32.8 & 41.0 & +8.2 \\
Streak Score
& 64.5 & 74.8 & +10.3
& 17.0 & 21.9 & +4.9
& 22.8 & 28.6 & +5.8
& 23.5 & 36.5 & +13.0
& 36.4 & 44.3 & +7.9
& 32.8 & 40.7 & +7.9 \\
Weighted Correctness
& 64.5 & 73.2 & +8.7
& 17.0 & 20.5 & +3.5
& 22.8 & 27.4 & +4.6
& 23.5 & 35.1 & +11.6
& 36.4 & 43.1 & +6.7
& 32.8 & 39.5 & +6.7 \\
First-Error Position
& 64.5 & 75.7 & +11.2
& 17.0 & 22.0 & +5.0
& 22.8 & 29.0 & +6.2
& 23.5 & 37.1 & +13.6
& 36.4 & 44.9 & +8.5
& 32.8 & 41.3 & +8.5 \\
\bottomrule
\end{tabular}%
}
\end{table*}

\begin{table*}[t]
\centering
\caption{\textbf{Best-of-8 reranking under five aggregation strategies, InternVL2.5-26B policy.}}
\label{tab:agg_26b}
\small
\setlength{\tabcolsep}{3.6pt}
\renewcommand{\arraystretch}{1.14}
\resizebox{\textwidth}{!}{%
\begin{tabular}{l ccc ccc ccc ccc ccc ccc}
\toprule
\rowcolor{headerblue}
& \multicolumn{3}{c}{\textbf{MathVista}}
& \multicolumn{3}{c}{\textbf{MathVision}}
& \multicolumn{3}{c}{\textbf{MathVerse-VO}}
& \multicolumn{3}{c}{\textbf{WeMath}}
& \multicolumn{3}{c}{\textbf{LogicVista}}
& \multicolumn{3}{c}{\textbf{Overall}} \\
\rowcolor{subheaderblue}
\textbf{Strategy}
& \textbf{P@1} & \textbf{B@8} & \textbf{$\Delta_8$}
& \textbf{P@1} & \textbf{B@8} & \textbf{$\Delta_8$}
& \textbf{P@1} & \textbf{B@8} & \textbf{$\Delta_8$}
& \textbf{P@1} & \textbf{B@8} & \textbf{$\Delta_8$}
& \textbf{P@1} & \textbf{B@8} & \textbf{$\Delta_8$}
& \textbf{P@1} & \textbf{B@8} & \textbf{$\Delta_8$} \\
\midrule
\rowcolor{bestrowgreen}
Geometric Mean
& 68.2 & \textbf{79.6} & \aggimp{+11.4}
& 23.4 & \textbf{28.1} & \aggimp{+4.7}
& 24.0 & \textbf{32.5} & \aggimp{+8.5}
& 30.9 & \textbf{42.1} & \aggimp{+11.2}
& 39.6 & \textbf{51.7} & \aggimp{+12.1}
& 37.2 & \textbf{46.8} & \aggimp{+9.6} \\
Correctness Rate
& 68.2 & 78.4 & +10.2
& 23.4 & 27.5 & +4.1
& 24.0 & 31.8 & +7.8
& 30.9 & 41.3 & +10.4
& 39.6 & 50.9 & +11.3
& 37.2 & 45.8 & +8.6 \\
Streak Score
& 68.2 & 78.0 & +9.8
& 23.4 & 27.2 & +3.8
& 24.0 & 31.4 & +7.4
& 30.9 & 41.0 & +10.1
& 39.6 & 50.5 & +10.9
& 37.2 & 45.4 & +8.2 \\
Weighted Correctness
& 68.2 & 76.5 & +8.3
& 23.4 & 26.0 & +2.6
& 24.0 & 30.1 & +6.1
& 30.9 & 39.6 & +8.7
& 39.6 & 49.1 & +9.5
& 37.2 & 44.0 & +6.8 \\
First-Error Position
& 68.2 & 79.0 & +10.8
& 23.4 & 27.9 & +4.5
& 24.0 & 32.2 & +8.2
& 30.9 & 41.7 & +10.8
& 39.6 & 51.2 & +11.6
& 37.2 & 46.3 & +9.1 \\
\bottomrule
\end{tabular}%
}
\end{table*}

\begin{table*}[t]
\centering
\caption{\textbf{Best-of-8 reranking under five aggregation strategies, InternVL2.5-38B policy.}}
\label{tab:agg_38b}
\small
\setlength{\tabcolsep}{3.6pt}
\renewcommand{\arraystretch}{1.14}
\resizebox{\textwidth}{!}{%
\begin{tabular}{l ccc ccc ccc ccc ccc ccc}
\toprule
\rowcolor{headerblue}
& \multicolumn{3}{c}{\textbf{MathVista}}
& \multicolumn{3}{c}{\textbf{MathVision}}
& \multicolumn{3}{c}{\textbf{MathVerse-VO}}
& \multicolumn{3}{c}{\textbf{WeMath}}
& \multicolumn{3}{c}{\textbf{LogicVista}}
& \multicolumn{3}{c}{\textbf{Overall}} \\
\rowcolor{subheaderblue}
\textbf{Strategy}
& \textbf{P@1} & \textbf{B@8} & \textbf{$\Delta_8$}
& \textbf{P@1} & \textbf{B@8} & \textbf{$\Delta_8$}
& \textbf{P@1} & \textbf{B@8} & \textbf{$\Delta_8$}
& \textbf{P@1} & \textbf{B@8} & \textbf{$\Delta_8$}
& \textbf{P@1} & \textbf{B@8} & \textbf{$\Delta_8$}
& \textbf{P@1} & \textbf{B@8} & \textbf{$\Delta_8$} \\
\midrule
\rowcolor{bestrowgreen}
Geometric Mean
& 71.9 & \textbf{83.5} & \aggimp{+11.6}
& 32.2 & \textbf{37.6} & \aggimp{+5.4}
& 36.9 & \textbf{47.7} & \aggimp{+10.8}
& 38.3 & \textbf{50.0} & \aggimp{+11.7}
& 47.9 & \textbf{58.7} & \aggimp{+10.8}
& 45.4 & \textbf{55.2} & \aggimp{+9.8} \\
Correctness Rate
& 71.9 & 82.3 & +10.4
& 32.2 & 36.8 & +4.6
& 36.9 & 46.8 & +9.9
& 38.3 & 49.1 & +10.8
& 47.9 & 57.8 & +9.9
& 45.4 & 54.3 & +8.9 \\
Streak Score
& 71.9 & 81.9 & +10.0
& 32.2 & 36.4 & +4.2
& 36.9 & 46.5 & +9.6
& 38.3 & 48.8 & +10.5
& 47.9 & 57.5 & +9.6
& 45.4 & 54.0 & +8.6 \\
Weighted Correctness
& 71.9 & 80.4 & +8.5
& 32.2 & 35.0 & +2.8
& 36.9 & 45.1 & +8.2
& 38.3 & 47.4 & +9.1
& 47.9 & 56.1 & +8.2
& 45.4 & 52.5 & +7.1 \\
First-Error Position
& 71.9 & 83.0 & +11.1
& 32.2 & 37.3 & +5.1
& 36.9 & 47.3 & +10.4
& 38.3 & 49.6 & +11.3
& 47.9 & 58.3 & +10.4
& 45.4 & 54.7 & +9.3 \\
\bottomrule
\end{tabular}%
}
\end{table*}

\definecolor{headerblue}{HTML}{DCEAF7}      
\definecolor{groupblue}{HTML}{EEF3F8}       
\definecolor{overallgreen}{HTML}{E8F4EA}    
\definecolor{metricgray}{HTML}{F7F7F7}      
\definecolor{goodgreen}{HTML}{2E7D32}       
\definecolor{warnred}{HTML}{C62828}         

\begin{table*}[t]
\centering
\caption{\textbf{Checklist completeness audit.} Under-reporting (omissions) and
over-reporting (false positives) of \texttt{visualdependency} compared to an
independent lenient visual-necessity annotation.}
\label{tab:checklist_audit}
\small
\setlength{\tabcolsep}{4.8pt}
\renewcommand{\arraystretch}{1.14}
\resizebox{\textwidth}{!}{%
\begin{tabular}{@{}l c c c c c c c@{}}
\toprule
\rowcolor{headerblue}
\textbf{Split} &
\textbf{\#Units} &
\textbf{\#Steps} &
\textbf{\#ShouldVis} &
\textbf{\#ModelVis} &
\textbf{\#Omit} &
\textbf{OmissionRate} &
\textbf{Completeness} \\
\midrule

\rowcolor{groupblue}
\multicolumn{8}{@{}l}{\textbf{Overall}} \\
\rowcolor{overallgreen}
\textbf{TOTAL} & 415 & 1780 & 1305 & 1314 & 77 & \textcolor{warnred}{0.0590} & \textcolor{goodgreen}{\textbf{0.9410}} \\
\midrule

\rowcolor{groupblue}
\multicolumn{8}{@{}l}{\textbf{By policy file}} \\
InternVL2.5-8B  & 153 & 689 & 453 & 419 & 48 & \textcolor{warnred}{0.1060} & 0.8940 \\
InternVL2.5-26B & 159 & 666 & 507 & 518 & 25 & 0.0493 & 0.9507 \\
InternVL2.5-38B & 103 & 425 & 345 & 377 & 4  & 0.0116 & \textcolor{goodgreen}{0.9884} \\
\midrule

\rowcolor{groupblue}
\multicolumn{8}{@{}l}{\textbf{By dataset}} \\
MathVerse-VO & 196 & 858 & 591 & 616 & 26 & 0.0440 & 0.9560 \\
MathVision   & 96  & 442 & 340 & 322 & 36 & \textcolor{warnred}{0.1059} & 0.8941 \\
WeMath       & 66  & 279 & 203 & 203 & 13 & 0.0640 & 0.9360 \\
MathVista    & 34  & 110 & 91  & 92  & 0  & 0.0000 & \textcolor{goodgreen}{\textbf{1.0000}} \\
LogicVista   & 23  & 91  & 80  & 81  & 2  & 0.0250 & 0.9750 \\
\bottomrule
\end{tabular}%
}
\end{table*}

\section{Checklist Completeness Audit for \texttt{visualdependency}}
\label{app:checklist_completeness}

EVPV uses the policy-provided \texttt{visualdependency} field as a step-wise
visual checklist (Section~\ref{sec:method:checklist}), which determines whether
a step is treated as visually dependent ($\nu_t=1$) and thus subject to
reliability gating (Eq.~\ref{eq:gated_reward}). A natural concern is whether a
policy could under-report visual dependency (e.g., outputting \texttt{null})
for steps that in fact rely on the image.

\paragraph{Audit protocol.}
We perform a human-verified audit on 415 sampled solution traces (1,780 total
steps) drawn from the same benchmark family used in our downstream evaluation.
For each step, we compare the policy flag $\nu_t=\mathbb{I}[\texttt{visualdependency}\neq\texttt{null}]$
with an independent annotation $\nu_t^\star$ indicating whether the step
\emph{should require} visual information. The annotation follows a \emph{lenient
policy} that favors recall: if a step plausibly depends on reading values,
identifying objects/labels, or using diagram/table/plot relations, we mark
$\nu_t^\star=1$; only clearly image-independent steps are marked $\nu_t^\star=0$.

\paragraph{Metrics.}
We count \textbf{omissions} (under-reporting) and \textbf{false positives}
(over-reporting):
\[
\text{Omission}=\mathbb{I}[\nu_t^\star=1\wedge \nu_t=0],\qquad
\text{FalsePos}=\mathbb{I}[\nu_t^\star=0\wedge \nu_t=1].
\]
We report:
\[
\textbf{OmissionRate}=\frac{\sum \text{Omission}}{\sum \mathbb{I}[\nu_t^\star=1]},\quad
\textbf{FalsePosRate}=\frac{\sum \text{FalsePos}}{\sum \mathbb{I}[\nu_t=1]},\quad
\textbf{Completeness}=1-\textbf{OmissionRate}.
\]

\paragraph{Results and discussion.}
Table~\ref{tab:checklist_audit} shows that omissions are limited in practice.
Overall, we observe an omission rate of \textbf{5.9\%} (completeness
\textbf{94.1\%}) and a false-positive rate of \textbf{7.1\%}. The omission rate
decreases with stronger policies (e.g., 38B has 1.2\% omission), suggesting that
visual dependency declaration behaves primarily as an instruction-following
task rather than an adversarial objective. Importantly, EVPV separates the
checklist from reward prediction: the base step rewards are produced by an
independent judge/verifier, while \texttt{visualdependency} only controls
whether reliability gating is applied. This decoupling reduces incentives for
reward hacking via systematically misreporting visual dependency, and the audit
confirms that large-scale under-reporting is not observed.

\definecolor{headerblue}{HTML}{DCEAF7}      
\definecolor{groupblue}{HTML}{EEF3F8}       
\definecolor{overallgreen}{HTML}{E8F4EA}    
\definecolor{totalgreen}{HTML}{DDEFE0}      
\definecolor{metricblue}{HTML}{1E88E5}      
\definecolor{textgray}{HTML}{4F4F4F}        

\begin{table*}[t]
\centering
\small
\setlength{\tabcolsep}{7pt}
\renewcommand{\arraystretch}{1.14}
\caption{\textbf{Human-annotated fidelity of extracted constraints.}
Precision/Recall/F1 of constraints predicted by $E_\phi$, evaluated on 600 sampled
instances (534 valid), reported by constraint type and dataset.}
\label{tab:extractor_fidelity_prf}
\begin{tabular}{lrrrrrr}
\toprule
\rowcolor{headerblue}
\textbf{Type} & \textbf{TP} & \textbf{FP} & \textbf{FN} & \textbf{Prec} & \textbf{Rec} & \textbf{F1} \\
\midrule

\rowcolor{groupblue}
\multicolumn{7}{c}{\textbf{LogicVista}} \\
numeric   & 19   & 6   & 3   & 0.7600 & 0.8636 & 0.8085 \\
relation  & 23   & 11  & 12  & 0.6765 & 0.6571 & 0.6667 \\
structure & 40   & 4   & 4   & 0.9091 & 0.9091 & 0.9091 \\
\rowcolor{overallgreen}
\textbf{OVERALL} & \textbf{82} & \textbf{21} & \textbf{19} & \textbf{0.7961} & \textbf{0.8119} & \textbf{0.8039} \\
\midrule

\rowcolor{groupblue}
\multicolumn{7}{c}{\textbf{MMMU}} \\
numeric   & 107  & 19  & 12  & 0.8492 & 0.8992 & 0.8735 \\
relation  & 77   & 20  & 8   & 0.7938 & 0.9059 & 0.8462 \\
structure & 114  & 8   & 7   & 0.9344 & 0.9421 & 0.9383 \\
\rowcolor{overallgreen}
\textbf{OVERALL} & \textbf{298} & \textbf{47} & \textbf{27} & \textbf{0.8638} & \textbf{0.9169} & \textbf{0.8896} \\
\midrule

\rowcolor{groupblue}
\multicolumn{7}{c}{\textbf{MathVerse-VO}} \\
numeric   & 421  & 91  & 35  & 0.8223 & 0.9232 & 0.8698 \\
relation  & 423  & 152 & 54  & 0.7357 & 0.8868 & 0.8042 \\
structure & 536  & 77  & 21  & 0.8744 & 0.9623 & 0.9162 \\
\rowcolor{overallgreen}
\textbf{OVERALL} & \textbf{1380} & \textbf{320} & \textbf{110} & \textbf{0.8118} & \textbf{0.9262} & \textbf{0.8652} \\
\midrule

\rowcolor{groupblue}
\multicolumn{7}{c}{\textbf{MathVision}} \\
numeric   & 193  & 43  & 23  & 0.8178 & 0.8935 & 0.8540 \\
relation  & 211  & 90  & 29  & 0.7010 & 0.8792 & 0.7800 \\
structure & 331  & 43  & 23  & 0.8850 & 0.9350 & 0.9093 \\
\rowcolor{overallgreen}
\textbf{OVERALL} & \textbf{735} & \textbf{176} & \textbf{75} & \textbf{0.8068} & \textbf{0.9074} & \textbf{0.8542} \\
\midrule

\rowcolor{groupblue}
\multicolumn{7}{c}{\textbf{MathVista}} \\
numeric   & 125  & 7   & 12  & 0.9470 & 0.9124 & 0.9294 \\
relation  & 86   & 19  & 11  & 0.8190 & 0.8866 & 0.8515 \\
structure & 132  & 10  & 7   & 0.9296 & 0.9496 & 0.9395 \\
\rowcolor{overallgreen}
\textbf{OVERALL} & \textbf{343} & \textbf{36} & \textbf{30} & \textbf{0.9050} & \textbf{0.9196} & \textbf{0.9122} \\
\midrule

\rowcolor{groupblue}
\multicolumn{7}{c}{\textbf{WeMath}} \\
numeric   & 102  & 29  & 18  & 0.7786 & 0.8500 & 0.8127 \\
relation  & 134  & 47  & 13  & 0.7403 & 0.9116 & 0.8171 \\
structure & 160  & 26  & 8   & 0.8602 & 0.9524 & 0.9040 \\
\rowcolor{overallgreen}
\textbf{OVERALL} & \textbf{396} & \textbf{102} & \textbf{39} & \textbf{0.7952} & \textbf{0.9103} & \textbf{0.8489} \\
\midrule

\rowcolor{totalgreen}
\multicolumn{7}{c}{\textbf{TOTAL}} \\
numeric   & 967  & 195 & 103 & 0.8322 & 0.9037 & 0.8665 \\
relation  & 954  & 339 & 127 & 0.7378 & 0.8825 & 0.8037 \\
structure & 1313 & 168 & 70  & 0.8866 & 0.9494 & 0.9169 \\
\rowcolor{totalgreen}
\textbf{OVERALL} & \textbf{3234} & \textbf{702} & \textbf{300} & \textbf{0.8216} & \textbf{0.9151} & \textbf{0.8659} \\
\bottomrule
\end{tabular}
\end{table*}

\section{Fidelity of the Structured Visual Constraints}
\label{app:extractor_fidelity}

EVPV relies on a constraint extractor $E_\phi$ to produce a reusable set of
structured visual facts $C=E_\phi(I,q)$ (Appendix~\ref{app:schema}). While these
constraints are trained via teacher distillation, it is important to verify
their faithfulness to the underlying images with a direct, human-annotated
evaluation.

\paragraph{Protocol.}
We randomly sample 600 instances from six multimodal reasoning benchmarks and
evaluate the predicted constraints against the images. After filtering
instances with missing/invalid outputs, 534 instances remain. For each instance,
annotators assess predicted constraints by type (\texttt{numeric},
\texttt{relation}, \texttt{structure}): a constraint supported by the image is
counted as a true positive (TP); an unsupported or contradictory constraint as a
false positive (FP); and a solution-critical visual fact missing from the
predicted set as a false negative (FN). We report Precision/Recall/F1:
$\mathrm{Prec}=TP/(TP+FP)$, $\mathrm{Rec}=TP/(TP+FN)$, and
$\mathrm{F1}=2\mathrm{Prec}\mathrm{Rec}/(\mathrm{Prec}+\mathrm{Rec})$.

\paragraph{Results.}
Table~\ref{tab:extractor_fidelity_prf} shows that the extractor achieves strong
overall fidelity: 0.8216 precision, 0.9151 recall, and 0.8659 F1.
By category, \texttt{structure} facts are the most reliable (0.9169 F1),
followed by \texttt{numeric} (0.8665 F1), while \texttt{relation} remains
the most challenging (0.8037 F1) due to fine-grained entity alignment (e.g.,
segment labels) and subtle geometric relations. Importantly, recall is high
across all categories, indicating broad coverage of solution-critical premises.

\paragraph{Discussion.}
Perfect constraint extraction is inherently difficult in visual math settings
(e.g., small text/OCR ambiguity, occlusion, and implicit or visually subtle
relations), and thus 100\% agreement is not expected. The results above
nevertheless indicate that $E_\phi$ produces high-quality, largely faithful
structured evidence suitable for EVPV: it provides both strong precision (to
avoid spurious support for hallucinated premises) and high recall (to cover the
facts required to validate grounded reasoning).

\section{Efficiency and Cost}\label{app:cost}
We compare EVPV with tool-integrated verification (TIM-PRM) from a deployment-cost perspective.
Table~\ref{tab:cost} summarizes the per-question inference cost in terms of (i) the number of extractor/judge/tool calls and (ii) a unified token/latency accounting.
EVPV performs a single constraint extraction per instance and reuses the resulting structured evidence across all steps (and candidates), thereby avoiding the per-step tool-execution loop used by TIM-PRM.
For TIM-PRM, we estimate the expected number of tool calls as $p_{\text{tool}}T$ using the tool-call frequency reported in the original paper, and highlight the resulting cost trade-off in the $\Delta$ row.

\begin{table}[htbp]
\centering
\caption{\textbf{Inference-time cost comparison (per question).}}
\label{tab:cost}
\small
\setlength{\tabcolsep}{8pt}
\renewcommand{\arraystretch}{1.3}

\begin{tabularx}{\linewidth}{@{}lCCC@{}}
\toprule
\rowcolor{PrimaryLight}
\textbf{Method} &
\textbf{Calls per question} &
\textbf{Expected cost formula} &
\textbf{vs. Baseline} \\
\midrule
\multicolumn{4}{@{}l}{\cellcolor{LightGray}\textit{Call breakdown (Extractor / Judge / Tool)}} \\
\addlinespace[2pt]

\textbf{EVPV (ours)} &
$1$ extractor + $T$ judge + $0$ tool &
$\text{Lat}(E_\phi) + T\!\cdot\!\text{Lat}(J)$ &
\cellcolor{SuccessLight}\textbf{Baseline} \\
\midrule

TIM-PRM~\cite{kuang2025tim} &
$0$ + $T$ + $p_{\text{tool}}T$ &
$T\!\cdot\!\text{Lat}(J) + p_{\text{tool}}T\!\cdot\!\text{Lat}(\text{tool})$ &
+$p_{\text{tool}}T$ tool calls \\
\midrule

\rowcolor{WarningLight}
\textbf{EVPV $-$ TIM-PRM} ($\Delta$) &
$+1$ extractor, $-\,p_{\text{tool}}T$ tools &
$\text{Lat}(E_\phi) - p_{\text{tool}}T\!\cdot\!\text{Lat}(\text{tool})$ &
\textbf{Saves} $p_{\text{tool}}T$ tools \\
\bottomrule
\end{tabularx}

\vspace{3mm}
\begin{minipage}{\linewidth}
\footnotesize
\textbf{Notation.} $T$: reasoning steps; $p_{\text{tool}}$: per-step tool probability ($\approx$0.21 from TIM-PRM paper); 
$\text{Lat}(\cdot)$: latency per call; EVPV extracts constraints \textit{once per instance} and reuses across all steps/candidates.
\end{minipage}
\end{table}


\section{Complete Ablation Results}
\label{app:ablations}

Table~\ref{tab:full_ablations} extends Table~4 of the main paper to include all
27 ablation configurations executed in Exp4. Configurations are organized by the
component being varied; the \emph{Full Method} row (EVPV + reliability gating)
is repeated at the top for reference. All scores are VisualProcessBench
Macro-F1 (\%); $\Delta$ is relative to the full method.

\noindent
Several additional observations emerge from Table~\ref{tab:full_ablations}.
First, history length shows a consistent monotonic trend: longer history is
better, but the marginal gain diminishes quickly beyond 4 steps, suggesting a
memory saturation effect. Second, vision sampling temperature has negligible
impact ($|\Delta|<0.5$), indicating robust constraint extraction under moderate
decoding variation. Third, parse-failure policy matters modestly ($|\Delta|\le 1.78$):
defaulting to $-1$ (conservative) slightly outperforms defaulting to $+1$ or random.

\definecolor{groupgray}{RGB}{245,245,245}  

\begin{table*}[t]
\centering
\caption{Complete ablation results on VisualProcessBench (Macro-F1, \%).
$\Delta$ = variant $-$ Full Method. Best per group is highlighted by the best score in that group (see group-wise shading).}
\label{tab:full_ablations}
\small
\setlength{\tabcolsep}{4.2pt}
\renewcommand{\arraystretch}{1.15}
\resizebox{\textwidth}{!}{%
\begin{tabular}{@{}l l cccccc c@{}}
\toprule
\textbf{Group} & \textbf{Variant}
& \textbf{DynaMath} & \textbf{MMMU}
& \textbf{MathVerse} & \textbf{MathVision}
& \textbf{WeMath} & \textbf{Overall} & $\boldsymbol{\Delta}$ \\
\midrule
\rowcolor{PrimaryLight}
\multicolumn{9}{@{}l}{\textbf{\color{PrimaryDark}Full Method (Reference)}} \\
Full Method
& \textbf{Full (EVPV + gating)}
& \textbf{69.57} & \textbf{68.86} & \textbf{67.09} & \textbf{65.27} & \textbf{69.11} & \textbf{67.46} & \textbf{+0.00} \\
\midrule

\rowcolor{LightGray}
\multicolumn{9}{@{}l}{\textbf{Evidence type}} \\
Evidence type
& w/o structured facts (caption-only)
& 67.75 & 58.09 & 63.48 & 60.68 & 67.10 & 63.38 & $-$4.08 \\
Evidence type
& w/o constraints (facts = $\emptyset$)
& 66.66 & 55.80 & 62.61 & 59.13 & 65.81 & 62.11 & $-$5.35 \\
Evidence type
& w/ shuffled facts (structure corrupted)
& 62.86 & 52.57 & 59.81 & 58.52 & 64.77 & 59.82 & $-$7.64 \\
Evidence type
& w/ noise caption only
& 64.41 & 56.22 & 61.05 & 59.80 & 65.33 & 61.18 & $-$6.28 \\
Evidence type
& Short vision prompt
& 68.02 & 66.14 & 65.73 & 63.91 & 67.44 & 66.05 & $-$1.41 \\
Evidence type
& w/ drop-facts corruption
& 34.90 & 34.40 & 36.29 & 36.14 & 35.96 & 35.77 & $-$31.69 \\
\midrule

\rowcolor{LightGray}
\multicolumn{9}{@{}l}{\textbf{Modality}} \\
Modality
& w/o vision (text-only judge, keep JSON)
& 58.44 & 49.44 & 53.59 & 54.07 & 61.02 & 54.93 & $-$12.53 \\
Modality
& w/o vision \& w/o JSON (text-only)
& 54.49 & 43.93 & 42.78 & 50.84 & 53.78 & 48.23 & $-$19.23 \\
Modality
& w/o vision JSON (keep image)
& 65.83 & 62.19 & 63.72 & 62.44 & 66.07 & 64.14 & $-$3.32 \\
\midrule

\rowcolor{LightGray}
\multicolumn{9}{@{}l}{\textbf{Judge prompt}} \\
Judge prompt
& Lenient judge prefix
& 66.91 & 65.28 & 64.02 & 62.75 & 67.09 & 65.13 & $-$2.33 \\
Judge prompt
& No-vision judge prefix
& 57.22 & 48.71 & 52.84 & 53.30 & 60.14 & 54.21 & $-$13.25 \\
Judge prompt
& Judge temperature 0.2
& 68.44 & 67.50 & 66.11 & 64.38 & 68.22 & 66.58 & $-$0.88 \\
Judge prompt
& Judge temperature 0.5
& 67.83 & 66.97 & 65.44 & 63.76 & 67.81 & 66.02 & $-$1.44 \\
\midrule

\rowcolor{LightGray}
\multicolumn{9}{@{}l}{\textbf{History length}} \\
History length
& History: none
& 65.74 & 63.21 & 62.80 & 61.45 & 65.53 & 63.49 & $-$3.97 \\
History length
& History: last 1 step
& 66.88 & 65.42 & 64.55 & 63.02 & 66.91 & 65.22 & $-$2.24 \\
History length
& History: last 2 steps
& 67.51 & 66.09 & 65.18 & 63.74 & 67.60 & 65.90 & $-$1.56 \\
History length
& History: last 4 steps
& 68.31 & 67.44 & 65.93 & 64.56 & 68.40 & 66.73 & $-$0.73 \\
History length
& History: last 8 steps
& 68.94 & 68.21 & 66.58 & 64.97 & 68.82 & 67.14 & $-$0.32 \\
\midrule

\rowcolor{LightGray}
\multicolumn{9}{@{}l}{\textbf{Vision decoding}} \\
Vision decoding
& Vision temperature 0.0
& 68.75 & 67.91 & 66.43 & 64.81 & 68.51 & 67.01 & $-$0.45 \\
Vision decoding
& Vision temperature 0.5
& 69.02 & 68.27 & 66.76 & 65.01 & 68.79 & 67.18 & $-$0.28 \\
Vision decoding
& Vision top-p 0.7
& 68.83 & 68.44 & 66.91 & 65.10 & 68.93 & 67.25 & $-$0.21 \\
\midrule

\rowcolor{LightGray}
\multicolumn{9}{@{}l}{\textbf{Parse-failure policy}} \\
Parse-failure
& Parse fail $\to +1$
& 67.44 & 66.31 & 65.02 & 63.19 & 67.25 & 65.68 & $-$1.78 \\
Parse-failure
& Parse fail $\to$ random
& 67.89 & 66.74 & 65.47 & 63.67 & 67.72 & 66.12 & $-$1.34 \\
Parse-failure
& Parse fail $\to -1$ (default)
& \textbf{69.57} & \textbf{68.86} & \textbf{67.09} & \textbf{65.27} & \textbf{69.11} & \textbf{67.46} & \textbf{+0.00} \\
\midrule

\rowcolor{LightGray}
\multicolumn{9}{@{}l}{\textbf{Compound ablations}} \\
Compound
& No vision JSON + text-only judge
& 53.11 & 42.87 & 41.64 & 49.72 & 52.45 & 47.07 & $-$20.39 \\
Compound
& Caption-only + no image in judge
& 56.72 & 47.39 & 49.81 & 52.14 & 57.03 & 52.49 & $-$14.97 \\
Compound
& Shuffled facts + lenient judge
& 61.45 & 50.88 & 57.93 & 56.71 & 62.24 & 57.94 & $-$9.52 \\
\bottomrule
\end{tabular}%
}
\end{table*}


\section{Qualitative Case Studies}
\label{app:cases}

We present three cases from VisualProcessBench. In each,
\textbf{process\_correctness} denotes the \emph{ground-truth} step-level labels
($+1$ = correct, $-1$ = incorrect). We show that EVPV-PRM's step-wise judgments
align with these labels by verifying the policy's visual claims against
extracted constraints $\mathcal{C}$.

\definecolor{EVPVBlue}{RGB}{55,130,180}
\definecolor{EVPVBlueLight}{RGB}{232,244,252}

\lstdefinestyle{evpvmono}{%
  basicstyle=\ttfamily\footnotesize,
  columns=fullflexible,
  breaklines=true,
  breakatwhitespace=false,
  showstringspaces=false,
  frame=none
}

\newtcolorbox{evpvcasepanel}[1]{%
  enhanced,
  colback=white,
  colframe=PrimaryBlue,
  boxrule=1.2pt,
  arc=4pt,
  outer arc=4pt,
  left=12pt,right=12pt,top=12pt,bottom=12pt,
  title={#1},
  coltitle=white,
  colbacktitle=PrimaryBlue,
  fonttitle=\bfseries\normalsize\sffamily,
  attach boxed title to top left={xshift=10mm,yshift=-4mm},
  boxed title style={
    sharp corners,
    boxrule=0pt,
    arc=3pt,
    left=10pt,right=10pt,
    top=3pt,bottom=3pt
  },
  shadow={3pt}{-3pt}{0pt}{black!25},
}

\newtcolorbox{evpvconstraintbox}{%
  enhanced,
  colback=PrimaryLight,
  colframe=PrimaryBlue!60,
  boxrule=0.8pt,
  arc=2pt,
  left=8pt,right=8pt,top=6pt,bottom=6pt,
  shadow={1pt}{-1pt}{0pt}{black!10},
}

\newcommand{\evpvsteptable}[6]{%
  \renewcommand{\arraystretch}{1.12}%
  \setlength{\tabcolsep}{5pt}%
  \begin{tabularx}{\linewidth}{@{}c X cc@{}}
    \toprule
    & \textbf{Step (abbrev.)} & \textbf{process\_correctness} & \textbf{EVPV step} \\
    \midrule
    #1\\
    #2\\
    #3\\
    #4\\
    #5\\
    #6\\
    \bottomrule
  \end{tabularx}%
}

\subsection{DynaMath: Misread kink position}
\label{app:case_fp}

\begin{figure*}[t]
\centering
\begin{minipage}{1\textwidth}
\begin{evpvcasepanel}{Case J.1: Graph --- continuous but not differentiable}

\textbf{Question (DynaMath):}
\textit{Determine for which values of $x=a$ the function is continuous but not differentiable at $x=a$.}
Gold answer: 1.

\vspace{6pt}
\centering
\includegraphics[width=0.62\textwidth]{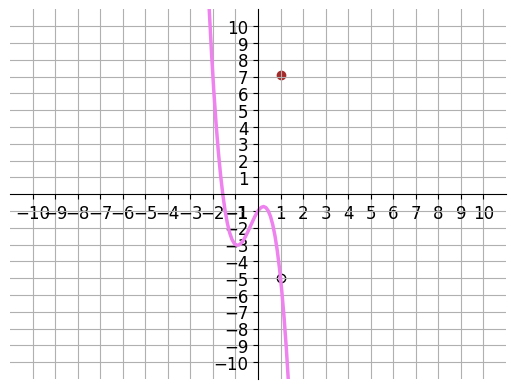}
\par\vspace{8pt}

\textbf{Extracted constraints $\mathcal{C}$ (by $E_\phi$):}
\begin{evpvconstraintbox}
\lstset{style=evpvmono}
\begin{lstlisting}
numeric:   {entity:"piecewise graph", attribute:"kink position", value:1, unit:"x"}
structure: {type:"graph", parts:["left branch","right branch"],
            attachment:["sharp corner at x = 1"]}
relation:  {type:"continuous_at", entities:["function","x=1"], confidence:0.95}
\end{lstlisting}
\end{evpvconstraintbox}

\vspace{4pt}
\textbf{Process-level verification.}
The policy claims a sharp corner at $x=-2$ (from step 3 onward); $\mathcal{C}$
gives the kink at $x=1$. Steps 3--6 thus contain an unsupported visual premise.
Matching yields low $p_j$ for those steps; reliability $r$ is attenuated and step
rewards are gated down.

\vspace{8pt}
\noindent
\evpvsteptable
{1 & Setup: find where continuous but not differentiable & $+1$ & $+1$}
{2 & Definitions (continuous / differentiable) & $+1$ & $+1$}
{3 & ``Sharp corner at $x=-2$'' (visual claim) & $-1$ & $-1$}
{4 & ``Therefore $x=-2$'' (conclusion) & $-1$ & $-1$}
{5 & ``The answer is $x=-2$'' & $-1$ & $-1$}
{6 & Verification of $x=-2$ & $-1$ & $-1$}

\vspace{6pt}
\noindent
EVPV-PRM matches the ground-truth process\_correctness: correct steps 1--2 are
preserved; incorrect steps 3--6 are flagged because the visual premise contradicts
$\mathcal{C}$.

\end{evpvcasepanel}
\end{minipage}
\end{figure*}

\subsection{MathVision: Unsupported geometric inference}
\label{app:case_fn}

\begin{figure*}[t]
\centering
\begin{minipage}{1\textwidth}
\begin{evpvcasepanel}{Case J.2: Quadrilateral angle (MathVision)}

\textbf{Question:}
\textit{In quadrilateral $ABCD$, $AD = BC$, $\angle \mathrm{DAC}=50^\circ$,
$\angle \mathrm{DCA}=65^\circ$, $\angle \mathrm{ACB}=70^\circ$.
How big is $\angle \mathrm{ABC}$?}
Gold answer: B ($55^\circ$).

\vspace{6pt}
\centering
\includegraphics[width=0.58\textwidth]{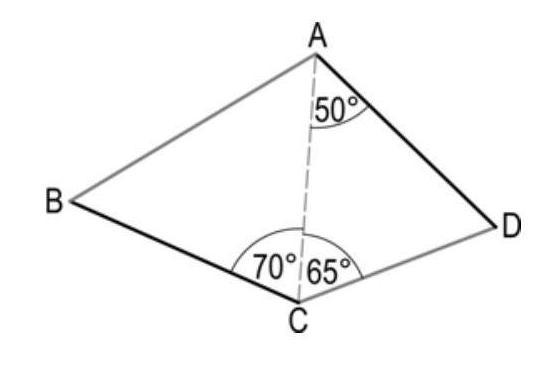}
\par\vspace{8pt}

\textbf{Extracted constraints $\mathcal{C}$ (by $E_\phi$):}
\begin{evpvconstraintbox}
\lstset{style=evpvmono}
\begin{lstlisting}
relation:  {type:"equal", entities:["AD","BC"], confidence:0.96}
numeric:   {entity:"angle DAC", value:50, unit:"deg"}, ...
structure: {type:"quadrilateral", parts:["A","B","C","D"]}
\end{lstlisting}
\end{evpvconstraintbox}

\vspace{4pt}
\textbf{Process-level verification.}
Step 1 only restates the problem and figure; its checklist items match $\mathcal{C}$.
Step 2 claims ``$\triangle ABC$ is isosceles with $AB=AC$'' from $AD=BC$; this claim
is not supported by $\mathcal{C}$ (equality is between $AD$ and $BC$, not $AB$ and $AC$).
Steps 2--5 are thus given low reliability and attenuated.

\vspace{8pt}
\noindent
\renewcommand{\arraystretch}{1.12}
\setlength{\tabcolsep}{5pt}
\begin{tabularx}{\linewidth}{@{}c X cc@{}}
\toprule
& \textbf{Step (abbrev.)} & \textbf{process\_correctness} & \textbf{EVPV step} \\
\midrule
1 & Task and given data ($AD=BC$, angles) & $+1$ & $+1$ \\
2 & $\triangle ACD$: derive an invalid angle statement & $-1$ & $-1$ \\
3 & ``$AD=BC \Rightarrow AB=AC$'', conclude $\angle\mathrm{ABC}=70^\circ$ & $-1$ & $-1$ \\
4 & Verify angles at $C$ & $-1$ & $-1$ \\
5 & Final answer D ($70^\circ$) & $-1$ & $-1$ \\
\bottomrule
\end{tabularx}

\vspace{6pt}
\noindent
Our method matches the ground truth: step 1 is correct and supported by $\mathcal{C}$;
steps 2--5 are incorrect and are correctly flagged because the key geometric premise
is unsupported.

\end{evpvcasepanel}
\end{minipage}
\end{figure*}

\subsection{WeMath: Mixed correct/incorrect steps, correct final answer}
\label{app:case_pi}

\begin{figure*}[t]
\centering
\begin{minipage}{1\textwidth}
\begin{evpvcasepanel}{Case J.3: Paper folding (WeMath)}

\textbf{Question:}
\textit{When the paper is folded with $\angle 1=\angle 2=\angle 3$, then $\angle 1$ equals ( ).
A. $90^\circ$ \; B. $45^\circ$ \; C. $60^\circ$ \; D. $30^\circ$ \; E. No correct answer.}
Gold answer: C ($60^\circ$).

\vspace{6pt}
\centering
\includegraphics[width=0.56\textwidth]{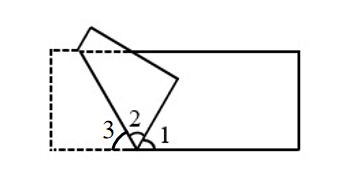}
\par\vspace{8pt}

\noindent
\begin{acmbluebox}{Extracted constraints $\mathcal{C}$ (by $E_\phi$)}
\footnotesize\ttfamily
\begin{verbatim}
[
  {
    "category": "relation",
    "type": "equal",
    "entities": ["angle 1", "angle 2", "angle 3"],
    "direction": null,
    "confidence": 0.90
  },
  {
    "category": "structure",
    "type": "sequence",
    "parts": ["fold line", "angle 1 region", "angle 2 region", "angle 3 region"],
    "attachment": ["angles are adjacent around the fold"],
    "adjacency": ["angle 1 adjacent to angle 2", "angle 2 adjacent to angle 3"],
    "confidence": 0.62
  }
]
\end{verbatim}
\end{acmbluebox}

\vspace{6pt}
\textbf{Process-level verification.}
The policy infers $60^\circ$ via ``angles form a triangle'' and ``equilateral''
(steps 2--3); the figure does not support that the three angles are interior
angles of one triangle. Steps 2--3 are incorrect; steps 4--6 (algebra and final
answer) are correct. EVPV assigns low $p_j$ to unsupported structural claims in
steps 2--3 and preserves reward for steps 4--6.

\vspace{8pt}
\noindent
\evpvsteptable
{1 & Key info: $\angle 1=\angle 2=\angle 3$ & $+1$ & $+1$}
{2 & ``Angles form a triangle; sum $180^\circ$'' & $-1$ & $-1$}
{3 & ``Equilateral; each $180^\circ/3$'' & $-1$ & $-1$}
{4 & ``Each angle $60^\circ$'' & $+1$ & $+1$}
{5 & ``Thus $\angle 1=60^\circ$'' & $+1$ & $+1$}
{6 & Final answer C & $+1$ & $+1$}

\vspace{6pt}
\noindent
EVPV-PRM's step-wise judgment matches process\_correctness exactly: incorrect
intermediate reasoning (steps 2--3) is flagged; correct conclusion steps (4--6)
are preserved.

\end{evpvcasepanel}
\end{minipage}
\end{figure*}

\end{document}

%% file: math_commands.tex
\usepackage{amsmath,amsfonts,bm}
\usepackage{natbib}
\usepackage{fancyhdr}

\def\eqref#1{equation~\ref{#1}}

\def\1{\bm{1}}

\DeclareMathAlphabet{\mathsfit}{\encodingdefault}{\sfdefault}{m}{sl}
\SetMathAlphabet{\mathsfit}{bold}{\encodingdefault}{\sfdefault}{bx}{n}

